%% file: root.tex
\documentclass[letterpaper, 10 pt, conference]{IEEEtran}  

\usepackage{graphics} 
\usepackage{epsfig} 
\usepackage{amsmath} 
\usepackage{amssymb}  

\usepackage[utf8]{inputenc} 
\usepackage[T1]{fontenc}    

\usepackage[dvipsnames]{xcolor}
\usepackage{booktabs}       
\usepackage{amsfonts}       
\usepackage{nicefrac}       
\usepackage{microtype}      

\usepackage{booktabs} 
\usepackage{multirow}
\usepackage{pifont}
\usepackage{setspace}
\usepackage{comment}
\usepackage{algorithm}
\usepackage[noend]{algpseudocode}

\usepackage{verbatim}
\usepackage{caption}
\usepackage{wrapfig}
\usepackage{svg}
\usepackage{mathabx}
\usepackage{subcaption}
\usepackage{enumitem}
\usepackage{color,soul}

\usepackage[noadjust]{cite}
\usepackage[english]{babel}
\usepackage{amsthm}
\theoremstyle{definition}

\theoremstyle{remark}

\setul{0.5ex}{0.3ex}
\definecolor{grey}{rgb}{.80,.80,0.80}
\setulcolor{grey}

\usepackage{hyperref}
\hypersetup{
    colorlinks=true,
    linkcolor=orange,
    filecolor=magenta,      
    urlcolor=orange,
    citecolor=orange,
}

\IEEEoverridecommandlockouts                              

\usepackage{amsfonts}               
\usepackage{amsmath}                
\usepackage{booktabs}               
\usepackage[T1]{fontenc}            
\usepackage{graphicx}               
\usepackage{hyperref}               
\usepackage{inconsolata}            
\usepackage[numbers,sort&compress]{natbib}        
\usepackage{nicefrac}               
\usepackage{xcolor}                 

\title{\LARGE \bf
Masked Imitation Learning: Discovering Environment-Invariant Modalities in Multimodal Demonstrations
}

\author{Yilun Hao$^{*}$,  Ruinan Wang$^{*}$, Zhangjie Cao, Zihan Wang, Yuchen Cui, Dorsa Sadigh
\thanks{*Equal contribution}
}

\begin{document}

\maketitle
\thispagestyle{empty}
\pagestyle{empty}

\begin{abstract}
Multimodal demonstrations provide robots with an abundance of information to make sense of the world. However, such abundance may not always lead to good performance when it comes to learning sensorimotor control policies from human demonstrations.
Extraneous data modalities can lead to state over-specification, where the state contains modalities that are not only useless for decision-making but also can change data distribution across environments. State over-specification leads to issues such as the learned policy not generalizing outside of the training data distribution.
In this work, we propose \emph{Masked Imitation Learning (MIL)} to address state over-specification by selectively using informative modalities. Specifically, we design a masked policy network with a binary mask to block certain modalities. We develop a bi-level optimization algorithm that learns this mask to accurately filter over-specified modalities. We demonstrate empirically that MIL outperforms baseline algorithms in simulated domains and effectively recovers the environment-invariant modalities on a multimodal dataset collected on a real robot. Our project website presents supplemental details and videos of our results at: {\color{RedOrange} https://tinyurl.com/masked-il}

\end{abstract}


\input{sections/1.intro}

\input{sections/2.related_work}

\input{sections/3.method}
\input{sections/4.experiments}

\input{sections/5.conclusion}

\bibliographystyle{IEEEtranN}
\bibliography{example}  

\clearpage

\input{sections/appendix.tex}

\end{document}

%% file: sections/1.intro.tex
\begin{figure*}[h]
    \centering
    \includegraphics[width=\textwidth]{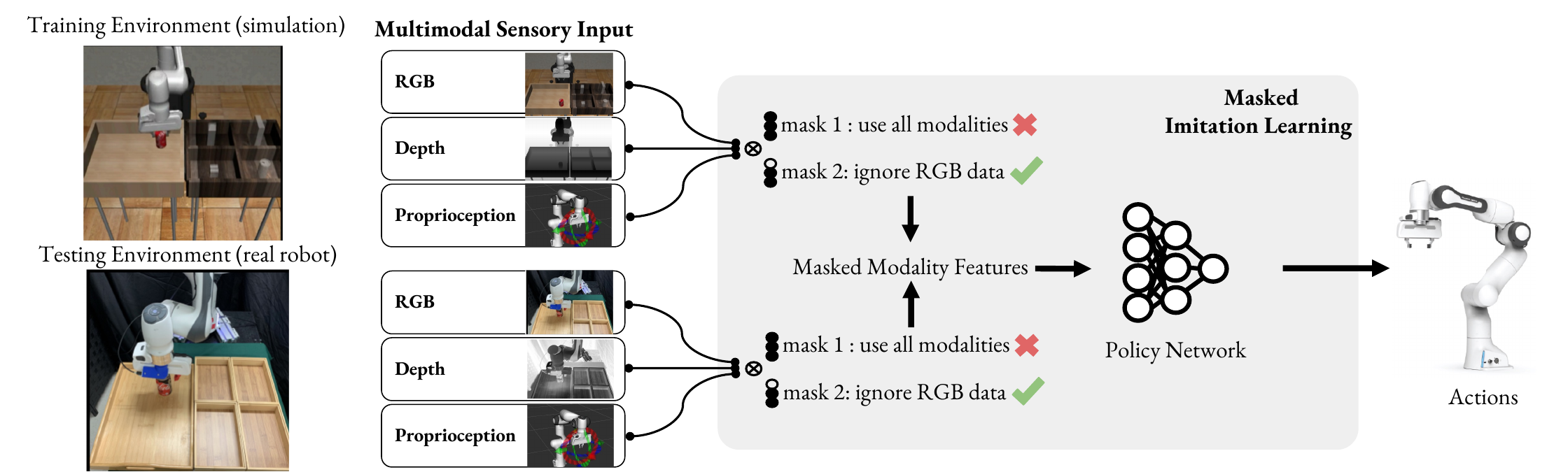}
    \vspace{-0.5cm}
    \caption{\small{An illustrative overview of masked imitation learning from multimodal data: we want to learn modality masks for selecting the input modalities that are invariant across different environments. In this example of transferring policies from sim to real, RGB modality has a larger domain gap than depth and proprioception. Masked imitation learning (MIL) would would learn to mask out the RGB modality because it induces lower loss on the real-world data using policies trained on sim data. }}
    \label{fig:overview}
    \vspace{-0.5cm}
\end{figure*}


\section{Introduction}
\label{sec:introduction}
Humans are born with the ability to perceive and integrate multiple sources of sensory information, including vision, touch, sound, and proprioception. 
Cognitive science research has demonstrated that humans form a coherent and robust perception of the world by efficiently integrating multiple sensory information~\cite{ernst2004merging,stein1993merging}. 
A robust perception is the foundation for sensorimotor control.
To mimic human sensing, modern robotic systems are often equipped with multitudes of sensors.  
Such \textit{multimodal} sensory information is important for solving robotics tasks, e.g., integrating visual inputs and haptic data are shown to be necessary for many contact-rich tasks~\cite{calandra2018more,zhang2019leveraging}. Similarly, both 3D point cloud and proprioception are shown to be necessary for decision making when interacting with objects with complex shapes~\cite{mu2021maniskill}.

Multimodal sensory data provides abundant information for decision making and much research effort has been dedicated to developing better robot sensing capabilities~\cite{liu2019flexible,abad2020visuotactile,dai2022design}. 
\textbf{However, somewhat unintuitively, more data modalities do not necessarily promise higher performance for sensorimotor policies learned from such data.}
For example, \citet{mandlekar2021matters} observed that when learning from multimodal proprioception data, using robot joint information and end-effector velocities in addition to end-effector pose leads to an inferior task success rate. Similarly, \citet{xiao2020multimodal} showed that in end-to-end driving, using only depth image can result in better performance than using both RGB and depth image when there is distributional shift in testing scenarios.
\citet{tomar2021model} also demonstrated that using all the data modalities gets higher loss than using a subset of modalities when learning a dynamics model for locomotion.
This phenomenon is caused by \textit{state over-specification}---the state contains extraneous data modalities that do not provide useful information for solving the task but can introduce a different data distribution across environments.
Consider the example in Fig.~\ref{fig:overview}, a robot is trained to perform a pick-and-place task in simulation from three different modalities (RGB image, depth image, and proprioception). To locate and pick up the object, depth and proprioception information are sufficient. The remaining modality of RGB image that changes from simulation to the real setting over-specifies the state for this particular task.
When such extraneous information is used by learned policies to predict action, they are less likely to generalize at test time, especially when the testing environment changes, e.g., when training in simulation and testing the policy on the real robot (see Fig.~\ref{fig:overview}). 
Prior work in learning from multimodal data only focuses on fusing various sensory modalities~\cite{akinola2020learning,liu2017learning,mu2021maniskill,lee2020making,lee2021detect,xiao2020multimodal}, but do not explicitly address the potential state over-specification problem.
In this paper, we focus on addressing this problem to avoid overfitting to training data when learning from multimodal data.
Going back to the example in Fig.~\ref{fig:overview}, 
if we remove the RGB modality, the robot still has sufficient information for performing the task and now the testing input looks more aligned with training data such that policy trained in simulation may generalize to the real environment.
Our key insight is that, for a particular task, we need to be \textit{selective} about what modalities the policy relies on for decision-making to avoid state over-specification.

We propose Masked Imitation Learning~(\textbf{MIL}), which learns a binary mask for each modality to decide whether the modality should be used for action prediction. 
MIL is a bi-level optimization algorithm: in the inner-level, MIL learns a policy embedded with a fixed mask on the training datasets using the training loss; in the outer-level, MIL updates the mask according to the validation loss evaluated on the validation dataset for the policy learned for each mask in the inner loop. The validation loss selects a policy that generalizes well to the validation dataset, which is more likely to be learned with no over-specified modalities, i.e. the corresponding mask removes over-specified modalities. We demonstrate the effectiveness of MIL empirically on several robotic tasks both in simulation including existing robotics datasets such as Robomimic~\cite{mandlekar2021matters} and on multimodal data collected for a real robot manipulation task.
We show learning from a selected set of modalities can improve the performance by 5.6$\times$ than learning from all modalities in certain domain.

%% file: sections/2.related_work.tex

\section{Related Work}

Our work addresses the state over-specification problem experienced by imitation learning from multimodal sensory data and therefore 
is closely related to prior work in \textit{learning from multimodal sensory data}, \textit{imitation learning}, and the broad area of \textit{invariant representation learning}. 

\medskip

\noindent \textbf{Learning from Multimodal Sensory Data.} 
Motivated by the potential of leveraging information from multiple sensory modalities, prior works have explored using multimodal data for robot learning.
These works mainly focus on what data modalities should be included for robot learning, where single-view images~\cite{lee2020making,calandra2018more,gao2016deep}, multi-view images~\cite{akinola2020learning}, haptic data~\cite{bekiroglu2011learning,calandra2018more,gao2016deep,sinapov2014learning}, range sensing~\cite{liu2017learning,sung2017learning}, audio~\cite{du2022play,zhang2019leveraging,dean2020see}, depth images~\cite{lee2020making} and 3D point clouds~\cite{mu2021maniskill} are adopted to learn different manipulation and navigation tasks. 
These works demonstrate that, for a given task, one can leverage a comprehensive set of modalities to provide the necessary information for decision-making.
Prior works have also investigated how to learn a robust state representation from multimodal sensory data using auxiliary objectives for better test-time generalization~\cite{yang2017deep,chen2021multi,lee2020making,lee2021detect}. 
Finally, a growing body of work also focuses on using multimodal data to learn end-to-end sensorimotor policies~\cite{ernst2004merging,liu2017learning,lee2020making,xiao2020multimodal,akinola2020learning,mu2021maniskill,du2022play}.
However, all these works utilize all the available modalities and do not consider the state over-specification problem that can occur due to extraneous modalities. 
We show that overfitting --- while being overlooked by most prior works --- is a common problem when learning from multimodal data, and we address it by our masked imitation learning approach. 

\medskip

\noindent \textbf{Imitation Learning.} 
Imitation learning aims to learn a policy from demonstrations~\cite{argall2009survey,osa2018algorithmic}. Behavioral cloning~\cite{bain1995framework} is the simplest form of imitation learning that treats the problem as supervised learning but often suffers from compounding errors at test time since the test data is not independent and identically distributed (i.i.d.).
More advanced imitation learning techniques such as generative adversarial imitation learning simultaneously learn a discriminator and the policy, addressing overfitting by rolling out the learned policies and applying a discriminative loss on agent trajectories that are far from those in demonstrations~\cite{schroecker2017state,torabi2018generative,sun2019provably,fu2017learning}. 
Further, several recent methods explicitly address the distribution shift between the demonstrations and the imitation agent's policy rollouts~\cite{liu2019state,cao2021corl}.
However, rolling out agent policies is not only expensive but also often unsafe for real-world robotics applications. 
In contrast, our algorithm learns from offline data and does not need to iteratively interact with the environment, and at the same time can be easily adapted to learn the environment-invariant modality mask in an online manner.

\noindent \textbf{Invariant Representation Learning.} 
Deep neural networks are known to have the capacity to memorize noise or pick up spurious correlations~\cite{zhang2021understanding,arpit2017closer}.
To reduce environment-specific overfitting, techniques including invariant risk minimization~\cite{arjovsky2019invariant,mahajan2021domain}, self-training~\cite{chen2020self,xie2020self}, dropout~\cite{srivastava2014dropout} and feature selection~\cite{yang1997comparative,bach2008bolasso,urbanowicz2018relief} are proposed to focus more on features with causal relationships to the outcome. However, all of these methods are only verified for non-robotics tasks.
For robotics tasks, several works propose information bottleneck to learn the task-relevant representation, which is invariant across domains~\cite{pacelli2020learning,lu2020dynamics}, but these approaches require a well-defined reward and interactions with the environment. Invariant risk minimization games is a theoretical framework to learn an invariant policy in many different environments to reduce the effect of spurious features~\cite{ahuja2020invariant}.
However, creating the set of environments that capture all the variations of spurious features is quite challenging especially in robotics domains. To address this, domain randomization approaches~\cite{tobin2017domain,peng2018sim,bousmalis2018using} try creating diverse environments by randomizing factors such as texture, lighting, etc., but a large number of variations of these factors, which often need to be done in simulation, might still not be able to capture all the spurious features. 
Inspired by the idea of invariant risk minimization that explicitly leverages the notion of \textit{environments}, our proposed method learns to mask out extraneous modalities that lead to poor generalization error in the validation environment so that our learned policies do not suffer from overfitting.





%% file: sections/3.method.tex
\section{Problem Setting}
\label{sec:problem}

We consider sequential decision-making problems modeled as Markov Decision Processes (MDPs). 
An MDP is defined by the tuple  $\langle \mathcal{S,A,T,R} \rangle$, where:
    $\mathcal{S}$ and $\mathcal{A}$ are the state space and action space; 
    $\mathcal{T}: \mathcal{S} \times \mathcal{A} \rightarrow \mathcal{S} $ is a transition probability function; and
     $\mathcal{R}$ is a reward function. 
Here, we focus on deterministic MDPs but as we discuss in the Appendix, our method can be easily extended to the stochastic case.
In this paper, we focus on the setting where the state space $\mathcal{S}$ consists of $M$ modalities: $\mathcal{S}=\bigtimes_{i=1}^{M}\mathcal{S}^i$, where each $\mathcal{S}^i$ indicates the state space of a modality.
A trajectory 
$\tau = \{(s_0,a_0),(s_1,a_1),...,(s_n,a_n)\}$ is a sequence of state-action pairs, where every state at time $t$ has $M$ modalities $s_t = [s^0_t,...,s^M_t]$.
The return of a trajectory is the sum of rewards $\sum_{t=0}^n[\mathcal{R}(s^i_t|_{i=1}^{M},a_t)]$.
Let $e \in \mathcal{E}$ denote an environment, which we define as a subspace of states $\mathcal{S}_e \subset \mathcal{S}$ that are reachable by the transition function $\mathcal{T}$ when initialized at $s_0 \in \mathcal{S}_e$.
A policy is a mapping from states to actions, $\pi: \mathcal{S} \rightarrow \mathcal{A}$. Similarly, an expert policy in an environment $e$ is $\pi^E: \mathcal{S}_e \rightarrow \mathcal{A}$ that maximizes the expected return. Expert demonstrations can be sampled from this policy to create a dataset: $\mathcal{D}_e =\{\tau_0,\tau_1,...,\tau_k\}$ of size $|\mathcal{D}_e|$.

The goal of \textbf{imitation learning} is to learn a policy generalizable across environments from multimodal demonstrations. 
Specifically, we have a training environment $e_{\text{train}}$, a validation environment $e_{\text{val}}$, and a testing environment $e_{\text{test}}$. 
Given a training dataset $\mathcal{D}_{\text{train}}$ with $|\mathcal{D}_{\text{train}}|$ demonstrations collected from $e_{\text{train}}$ and a validation dataset $\mathcal{D}_{\text{val}}$ with $|\mathcal{D}_{\text{val}}|$ demonstrations collected from $e_{\text{val}}$, our goal is to learn a policy that achieves high performance in $e_{\text{test}}$ using the data in $\mathcal{D}_{\text{train}}$ and $\mathcal{D}_{\text{val}}$ ($|\mathcal{D}_{\text{train}}|\gg |\mathcal{D}_{\text{val}}|$). 
We assume that, when providing demonstrations, the expert has access to the same raw state, hence the same data modalities, as the learning agent does.


Imitation learning algorithms often learn a policy that minimizes the training loss $\mathcal{L}$ on the training dataset 
In practice, this loss function $\mathcal{L}$ is usually the maximum likelihood loss or L2 distance for continuous action spaces or a cross-entropy loss for discrete action spaces.
 During training, the best model is selected by validation loss, which is of the same form as the training loss but is evaluated on the validation dataset $\mathcal{D}_{\text{val}}$. \textbf{Overfitting} is a phenomenon that the model learned by the training loss on $\mathcal{D}_{\text{train}}$ and selected by the validation loss on $\mathcal{D}_{\text{val}}$ performs poorly on the test environment $e_{\text{test}}$. 

In this work, we focus on addressing overfitting caused by \textbf{state over-specification}, which happens when the state observed by the imitating agent contains more modalities than what was used by the demonstrator to perform the task, and such modalities change the data distribution across environments. For example, to perform the task of \textit{cutting an apple into slices}, a human demonstrator only needs the location, shape, and size of the apple while the texture of the cutting board or the color of the knife handle is useless and may change in different kitchens. Though both observing the full state information, a human demonstrator selects the useful and generalizable modalities in the state to make decisions, enabling humans to perform the task across environments, but an imitating agent may \textit{overfit} to the over-specified modalities, e.g., if we have only observed knives with green handles in training, the agent at test time can only cut the apple when the knife's handle is green.



\noindent \textbf{Problem Statement.}  Let $s^i|_{i=0}^M\in \mathcal{S}$ denote the full state with $M$ modalities and $s^i_*|_{i=0}^N\in \mathcal{S^*}$ denote the modalities of size $N$ ($N$$\leq$$M$), which the expert uses to make decisions ($\mathcal{S}^*$ is the environment-invariant modalities for the demonstrated task). 
We let $\bar{s}|_{i=0}^{M-N} \in \bar{\mathcal{S}}$ of size $M-N$ denote the extraneous modalities that are not used by the expert to act.
Our goal is to find a policy $\pi_\theta: \mathcal{S} \rightarrow \mathcal{A}$ trained and validated on $\mathcal{D}_\text{train}$ and $\mathcal{D}_\text{val}$ respectively, which matches the performance of an expert demonstrator $\pi^E: \mathcal{S^*} \rightarrow \mathcal{A}$ in test environment $e_\text{test}$, while the agent observes the full state including the extraneous modalities $\bar{s} \in \bar{\mathcal{S}}$. We define this problem as the state over-specification problem for imitation learning from multimodal data.

\section{Masked Imitation Learning}
\label{sec:method}


To address overfitting caused by state over-specification, our key insight is to be \textit{selective} about what modalities the policy relies on for deicision making, and remove the over-specified modalities and only preserve the modalities that are generalizable across environments. 
We develop a masked imitation learning (MIL) method to achieve this. 
In this section, we present the model architecture and discuss loss design.

\begin{figure*}[t]
    \centering
    \includegraphics[width=0.95\linewidth]{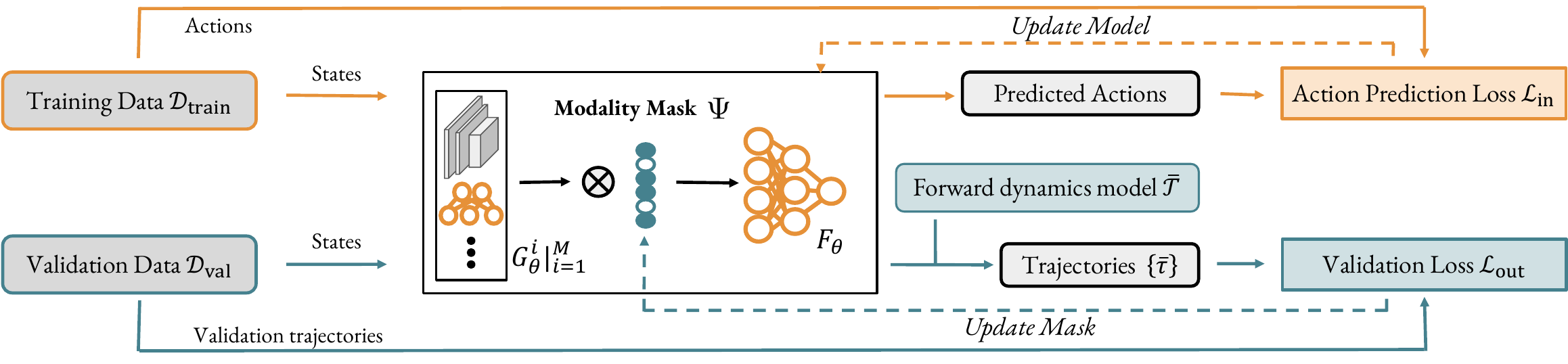}
    \caption{\small{Masked imitation learning (MIL) from multimodal data: the inner loop of MIL takes a fixed mask and uses standard behavioral cloning objective to optimize the policy network; and the outer loop of MIL employs a validation loss to update the binary modality mask.}}
    \label{fig:approach}
    \vspace{-0.3cm}
\end{figure*}

\subsection{Policy Network Architecture}
Fig.~\ref{fig:approach} shows an overview of our proposed masked imitation learning (MIL) method. 
Our model consists of three parts: the feature encoder for each modality $G_\theta^i|_{i=1}^{M}$, a learnable binary mask vector $\Psi \in \{0,1\}^M$ with one bit mask for each modality, and an action predictor $F_\theta$. 
We use $\theta$ to denote all the parameters of the encoders $G^i|_{i=1}^{M}$ and the action predictor $F$. 
The feature encoder $G^i$ extract a feature vector $G^i(s^i)$ from the $i$-th modality. Then the feature $G^i(s^i)$ is multiplied by the $i$-th dimension of the mask $\Psi[i]$ and all the masked features are then concatenated into a single feature vector: $[\Psi^1 G^1(s^1), \dots, \Psi^M G^M(s^M)]$, where $F$ uses this masked featurized state to predict the final action. 
In all, our policy can be represented as:
\begin{equation}
    \pi(s;\Psi,\theta) = F_\theta\left([\Psi[1] G_\theta^1(s^1), \dots, \Psi[M] G_\theta^M(s^M)]\right).
\end{equation}

\subsection{Bi-Level Optimization}
The goal of MIL is to simultaneously optimize $G_\theta^i|_{i=1}^{M}$ and $F_\theta$ and find a mask $\Psi$ that assigns zero weight to over-specified modalities. To achieve this, at the high level, we develop a bi-level optimization framework, where the inner-level takes a fixed mask and optimizes $G_\theta^i|_{i=1}^{M}$ and $F_\theta$ with imitation training loss $\mathcal{L}_\text{in}$ using standard gradient descent over $\theta$ and the outer loop optimizes the modality mask $\Psi$ with the validation loss $\mathcal{L}_\text{out}$ using the coordinate descent algorithm~\citep{wright2015coordinate}. 
The key idea of MIL is that the inner-level optimization process could find a model that minimizes the imitation learning loss on the training data for a specific mask, and the outer loop evaluates the generalizability of the learned model with a validation loss and decides whether the mask selects the robust modalities.

Specifically, following the coordinate descent algorithm that is widely used for learning binary variables, we start from a mask $\Psi$ with all the entries as one and iteratively update the mask from the first bit to the last bit cyclicly one bit at a time. 
At each iteration, we have a mask $\Psi$ and a bit index $j$ indicating which bit we want to update in this iteration. We create two masks $\Psi^0$ and $\Psi^1$ by setting the $j$-th bit of $\Psi$ as $0$ and $1$ respectively. We then execute the inner-level imitation learning process for $\Psi^0$ and $\Psi^1$ respectively with the inner-level imitation loss, where we take the widely-adopted L2 loss optimized on the training data $\mathcal{D}_\text{train}$:
\begin{equation}
    \mathcal{L}_\text{in}(\theta) = \mathbb{E}_{(s,a)\in \mathcal{D}_\text{train}}||\pi(s;\Psi,\theta)-a||^2.
\end{equation}
After convergence, we learn the parameters $\theta(\Psi)$ that optimize the imitation learning loss $\mathcal{L}_{\text{in}}$ for the given masks $\Psi^0$ and $\Psi^1$. In the outer loop, after deriving the parameters $\theta(\Psi^0)$ and $\theta(\Psi^1)$ that optimize $\mathcal{L}_{\text{in}}$ for $\Psi^0$ and $\Psi^1$ respectively, we evaluate the generalizability of both parameters with a validation loss with the same form as the training loss but on the validation data $\mathcal{D}_{\text{val}}$:
\begin{equation}\label{eqn:valid}
    \mathcal{L}_\text{out}(\Psi) = \mathbb{E}_{(s,a)\in \mathcal{D}_\text{val}}||\pi(s;\Psi,\theta(\Psi))-a||^2.
\end{equation}
The only difference between $\mathcal{L}_\text{in}$ and $\mathcal{L_\text{out}}$ is the dataset they are trained on, i.e., $\mathcal{D}_\text{train}$ vs. $\mathcal{D}_\text{val}$. The mask that includes  extraneous modalities will overfit to the training data and introduce a high $\mathcal{L}_\text{out}$ on $\mathcal{D}_\text{val}$. Thus, at the $j$-th iteration (corresponding to the $j$-th bit of the mask), we select the mask ($\Psi^0$ or $\Psi^1$) that minimizes the outer loss $\mathcal{L}_\text{out}$ to update the $j$-th bit of mask. We then repeat this procedure to update the next bit in the mask.

\begin{figure}[b]
\vspace{-0.5cm}
 \centering 
    \includegraphics[width=0.36\textwidth]{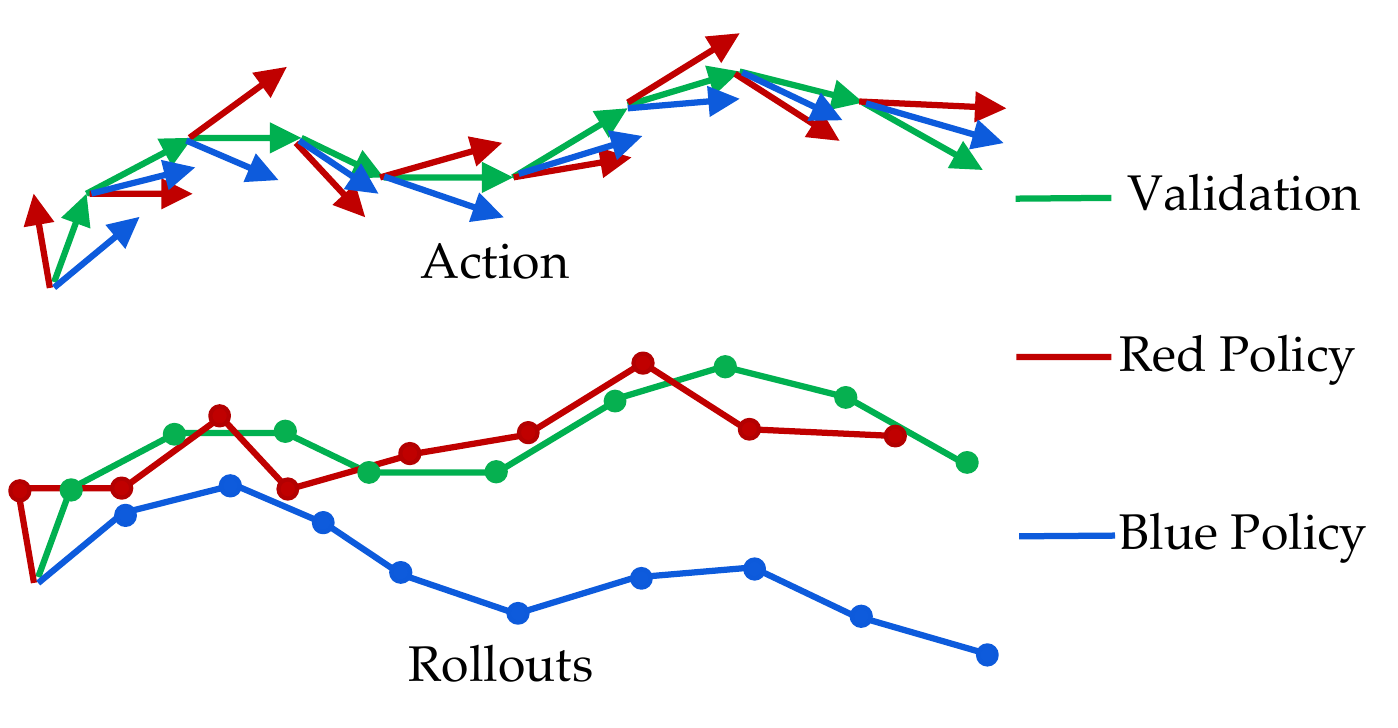}  
  \caption{\small{Motivation for using a state-based loss: Blue policy has a lower action loss compared to Red policy on a given set of states from the validation dataset, but its trajectory starting from the same initial state deviates more from the validation trajectory.}}
  \label{fig:state_loss}
\end{figure}

\noindent \textbf{Remark.} Note that although we adopt similar training and validation loss as common imitation learning algorithms, the modality mask and the bi-level optimization process allow MIL to learn a more generalizable policy. 
In a common imitation learning setting, the policy uses all the modalities and cannot avoid overfitting to extraneous modalities, the validation loss can only select the most generalizable policy within a pool of overfitting policies. However, with MIL, the validation loss selects which mask to use instead of selecting $\theta$, which allows it to remove extraneous modalities that cause overfitting leading to learning a more generalizable imitation policy. 

\subsection{Updating Validation Loss Using a Forward Dynamics Model}

\begin{figure*}[btp]
    \centering
    \begin{subfigure}[t]{.2\linewidth}
        \centering
        \includegraphics[width=\linewidth, height=\linewidth]{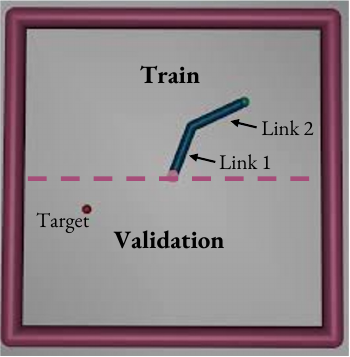}
        \caption{\texttt{Reacher}}
        \label{fig:reacher}
        \vspace{0.1cm}
    \end{subfigure}
    \hfil
    \begin{subfigure}[t]{0.2\linewidth}
        \centering
        \includegraphics[width=\linewidth, height=\linewidth]{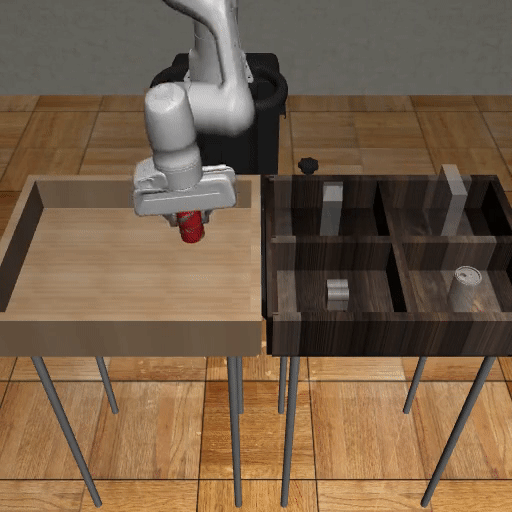}
        \caption{\texttt{Robomimic-Can}}
        \label{fig:robomimic-can}
    \end{subfigure}
    \hfil
    \begin{subfigure}[t]{0.2\linewidth}
        \centering
        \includegraphics[width=\linewidth, height=\linewidth]{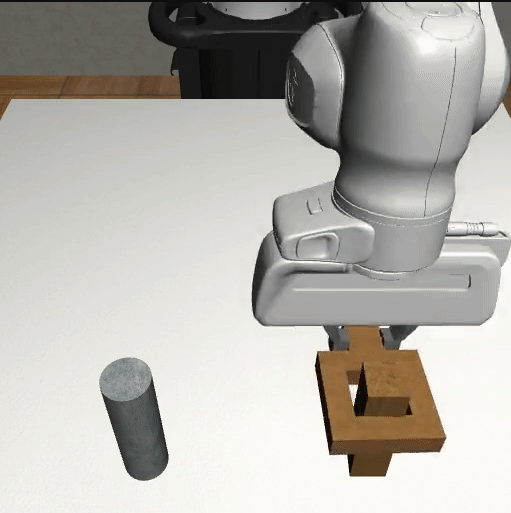}
        \caption{\texttt{Robomimic-Square}}
        \label{fig:robomimic-square}
    \end{subfigure}    
    \hfil
    \begin{subfigure}[t]{0.2\linewidth}
        \centering
        \includegraphics[width=\linewidth, height=\linewidth]{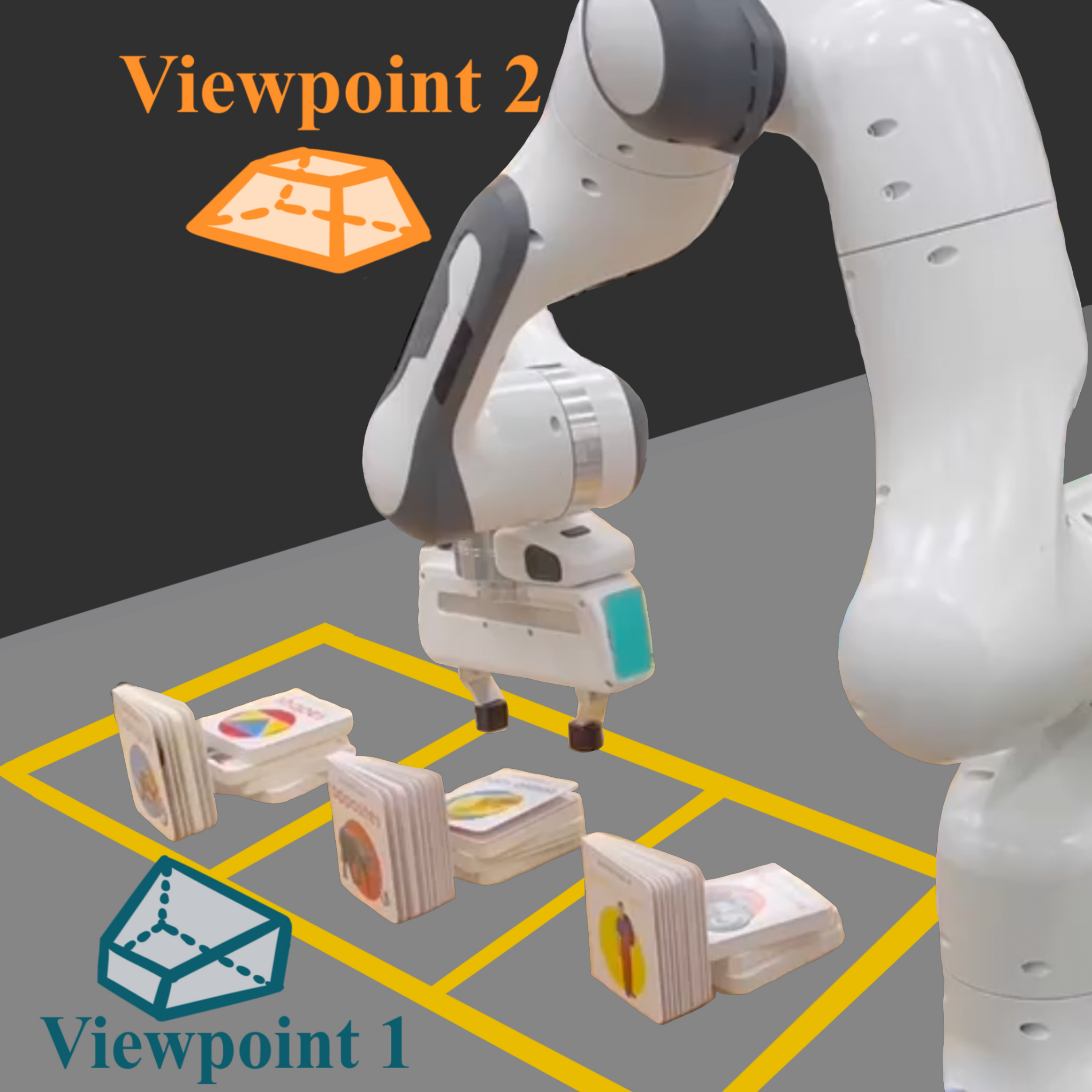}
        \caption{\texttt{Bookshelf} (real robot)}
        \label{fig:realrobot}
    \end{subfigure}
    \vspace{-0.2cm}
    \caption{Experimental task domains.}
    \label{fig:domains}
    \vspace{-0.4cm}
\end{figure*}

The current validation loss defined on state-action pairs in Eqn.~\eqref{eqn:valid} only evaluates the per-step error of actions. However, as demonstrated in prior works~\cite{ross2011reduction,ross2010efficient}, imitation learning suffers from large compounding errors across long sequences even though the error in each step can be small. As a concrete abstract example, in Fig.~\ref{fig:state_loss}, we have two policies (Red and Blue) each generating a sample trajectory shown in the figure. A sample trajectory from the validation dataset starting from the same initial state is also shown in green. Even though the Blue policy has lower action loss compared to the Red policy, its rollout deviates more from the validation trajectory. Therefore, the per-step validation loss sometimes cannot accurately evaluate the performance of a policy on the validation dataset, which may lead to selecting a suboptimal mask. 
Instead, we would like to use the distance between the policy rollouts and validation trajectories as the validation loss such that Red policy achieves a lower loss than the Blue policy.
Since our method is offline, we cannot generate rollouts by interacting with the environment. Instead, we learn a forward dynamics model $\overline{\mathcal{T}}$ to approximate the transition dynamics $\mathcal{T}$ based on $\mathcal{D}_\text{val}$. 
Let us define the forward dynamics loss as:
\begin{equation}
    \mathcal{L}(\overline{\mathcal{T}}) = \mathbb{E}_{(s_t,a_t,s_{t+1})\in \mathcal{D}_\text{val}}\lVert\overline{\mathcal{T}}( s_t,a_t)-s_{t+1}\rVert^2.
\end{equation}
Now, we can create a new trajectory $\bar{\tau}$ by rolling out the policy $\pi(s;\Psi,\theta)$ from an initial state $s_0$ of each trajectory $\tau$ in $\mathcal{D}_{\text{val}}$ with the forward dynamics model $\overline{\mathcal{T}}$ as follows:
\begin{equation}
    \bar{s}_0 = s_0, \quad \bar{s}_{t+1} = \overline{\mathcal{T}}(\bar{s}_t, \pi(\bar{s}_t;\Psi,\theta(\Psi))).
\end{equation}
By rolling out the policy from each initial state in $\mathcal{D}_{\text{val}}$ for the same length as the corresponding trajectory in the validation set, we create a new dataset $\bar{\mathcal{D}}_\text{val}$, within which each state $\bar{s}$ has a corresponding state $s$ in $\mathcal{D}_{\text{val}}$. 
We then compute the outer loop validation loss based on the state differences between $\bar{s}\in \bar{\mathcal{D}}_\text{val}$ and $s\in\mathcal{D}_\text{val}$:
\begin{equation}
    \mathcal{L}_\text{out}(\Psi) = \mathbb{E}_{(\bar{s},s)\in(\bar{\mathcal{D}}_\text{val},\mathcal{D}_\text{val})}\lVert \bar{s}-s\rVert^2.
\end{equation}
This new validation loss using the learned forward dynamics model is designed for tasks with long-horizon trajectories or tasks that consist of multiple stages, which can suffer from large compounding errors. Note that the forward dynamics model suffers less from compounding error because we only query data near the trajectories in the validation set. In practice, we train the forward dynamics model with the proprioception states to avoid having to learn visual dynamics models that can be much less accurate.

\noindent \textbf{Remark.} The key elements of MIL are a bi-level optimization that learns a binary mask in the outer loop using coordinate descent and a learned forward dynamics model for constructing validation loss of long-horizon tasks. 
The full algorithm is presented in Appendix A. 
Note that MIL learns to filter extraneous modalities that induce large generalization errors (high validation loss) but do not necessarily return the smallest number of modalities needed to learn a task. If there are redundant modalities that do not influence the performance of the learned policy (whether it was used by the expert or not), MIL may not learn to filter them. 

%% file: sections/4.experiments.tex
\begin{table*}
\caption{Success rates of policies learned with different methods.}
\label{table:succes-rates}
\centering
\renewcommand{\arraystretch}{1.08}
\small
\vspace{-0.1cm}
\begin{tabular}{c|cccccccc}
 \midrule 
 
\small{Task}              & \small{MIL (ours)}     & \small{MIL-aug (ours)}  & \small{MaskDropout} & \small{MaskAverage} & \small{BC-NoMask} & \small{OracleMask} & \small{ContinuousMask} \\ \midrule 
\texttt{RM-Can}     & \textbf{56.0$\pm$5.3}  & -   & 35.3$\pm$30.6   & 30.7$\pm$8.6    & 22.7$\pm$29.5 & \textbf{56.0$\pm$5.3}   & 47.3$\pm$4.16   \\
\texttt{RM-Square} & 56.7$\pm$9.8 & \textbf{71.3$\pm$4.7} & 19.3$\pm$4.5    & 18.1$\pm$14.3   & 12.7$\pm$5.2  & 59.3$\pm$8.2   & 2.7$\pm$1.2      \\
\texttt{Bookshelf}        & \textbf{95.24}    & -        & 47.9        & 60.9        & 54.17     & \textbf{95.24}      & 19.8   \\  \midrule 
\end{tabular}
\vspace{-0.2cm}
\end{table*}

\begin{figure*}[]
\vspace{-0.1cm}
    \centering
    \begin{subfigure}[t]{0.3\linewidth}
    \centering
    \includegraphics[width=0.9\linewidth]{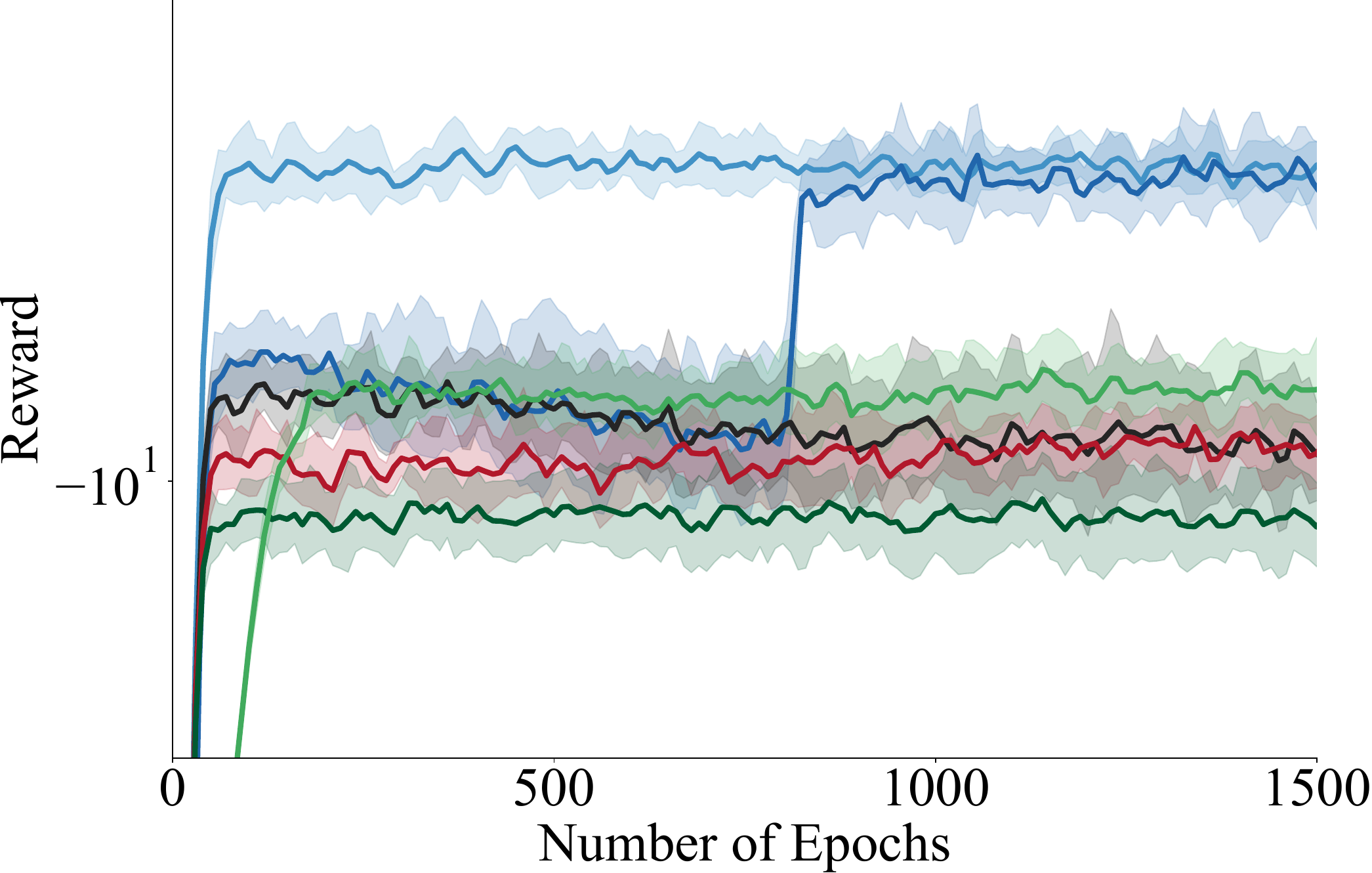}
    \caption{\small{Distributionally Shifted Goals }}
    \label{fig:my_label}
    \end{subfigure}
    \hfil
    \begin{subfigure}[t]{0.3\linewidth}
    \centering
    \includegraphics[width=0.9\linewidth]{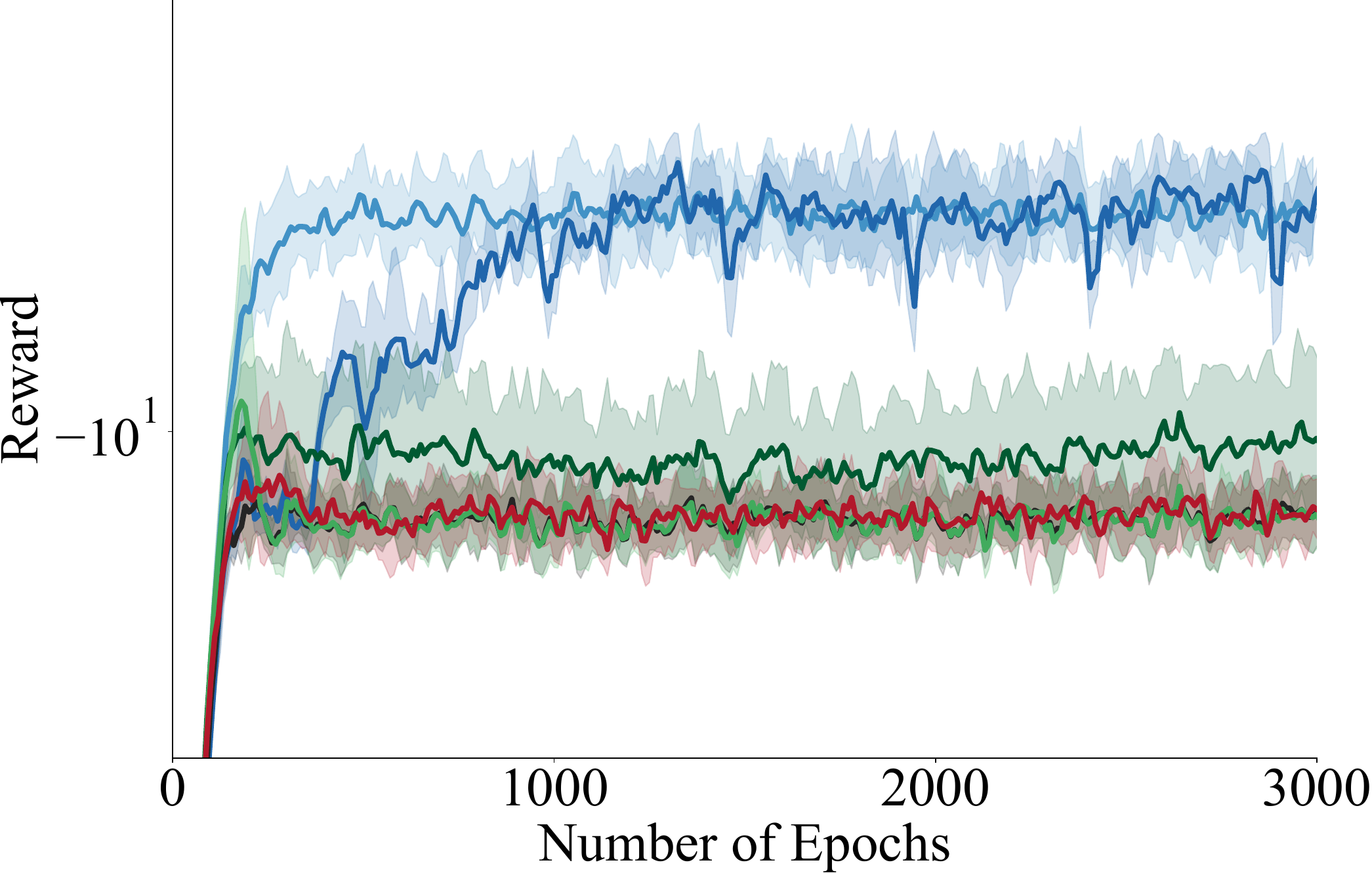}
    \caption{\small{Using Image Observations}}
    \label{fig:my_label}
    \end{subfigure}
    \begin{subfigure}[t]{0.15\linewidth}
    \includegraphics[width=\linewidth]{figures/legend2.pdf}
    \end{subfigure}
    \vspace{-0.1cm}
    \caption{Performance (reward) of learned policies in \texttt{Reacher}}
    \label{fig:reacher_results}
    \vspace{-0.4cm}
\end{figure*}

\section{Experiments}
\label{sec:experiments}

We evaluate MIL for imitation learning in two simulated robotic control tasks and on a multimodal dataset collected on a Franka Panda arm. 
We compare MIL with several baselines and ablations including BC-NoMask (vanilla BC using all the modalities), MaskDropout (the policy is learned with random dropout on the mask), MaskAverage (average performance of policies learned with randomly selected but fixed masks), OracleMask (a manually selected mask by oracle/designer), ContinuousMask (a continuous/non-binary mask is learned altogether with the policy net using SGD), and MIL with online evaluation (allowing interactions with the validation environment).

\medskip

\textbf{Reacher} is a MuJoCo-based task where an agent with a two-link arm needs to reach a specified target location in 2 dimensional space. We create two experiment settings in this environment. 
1) In the first setting, as shown in Fig.~\ref{fig:reacher}, we divided the 2D space into 2 different regions for sampling the target such that train and validation/testing environments have their own target distribution. \ul{The 5 input modalities are the angle between two links, the angle between the first link and the target, the distance between the target and the center, the angular velocities of the first and second links, and the target position in the Cartesian coordinate system}. Here, we expect that the model may overfit to the target position in the Cartesian coordinate system, which has a distribution shift from the top half to the bottom half.
2) In the second setting, \ul{the 4 input modalities include the rendered RGB image and the low-dimensional states in Reacher (the angular velocities of the first and second links, the target position and the relative location of the fingertip and the target).}

The results in this task domain are presented in Fig.~\ref{fig:reacher_results}, in which MIL outperforms baseline methods in both settings. MIL reaches a higher reward once it learned to mask out the extraneous modalities (the Cartesian target position for the first setting and the image for the second setting). 
In this task, we also verify that \textit{redundant modalities} --- modalities containing the exact same information as existing ones or that can be derived from existing ones --- do not influence the performance of the learned policy.

\begin{figure}[h]
  \centering
   \includegraphics[width=0.85\linewidth]{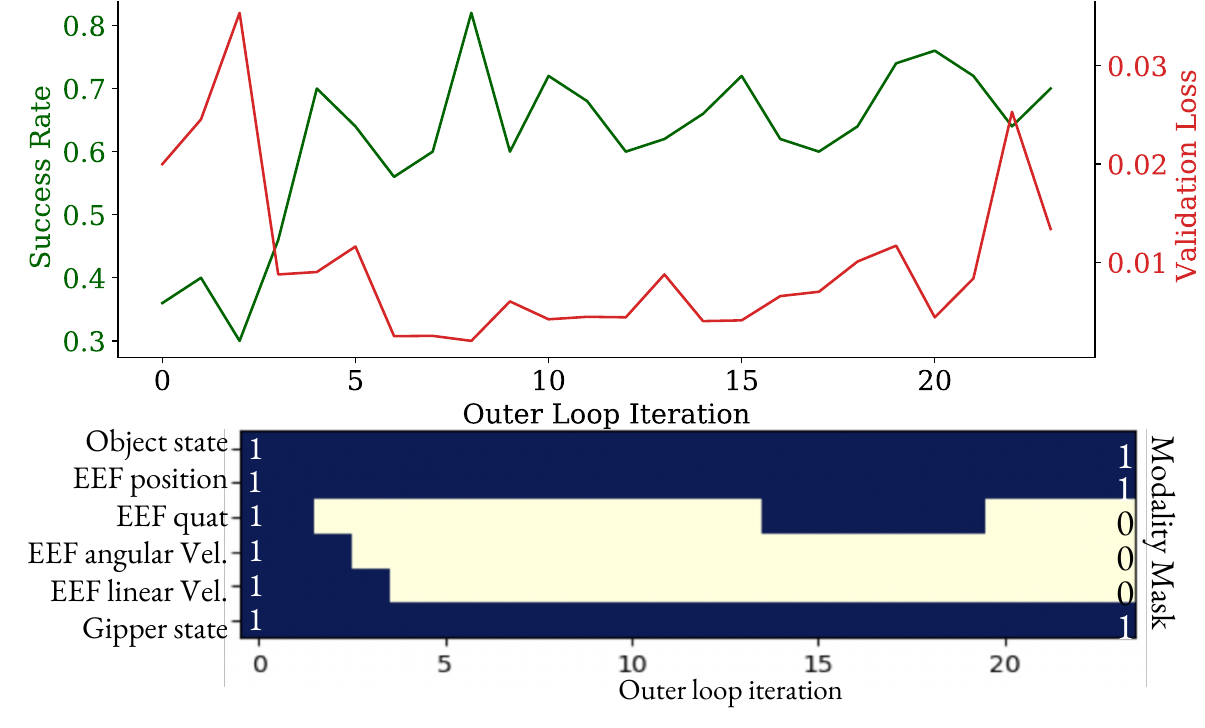}
        \caption{\small{Example run of MIL: mask updates over outer loops with the validation loss and success rate.}}
        \label{fig:robomimic_sample}
  \vspace{-0.4cm}
\end{figure}

\medskip

\textbf{Robomimic-Can} is a simulated task from the work of Mandlekar et al.~\cite{mandlekar2021matters}, where a Franka Panda robot is learning to pick up a can and put it inside a target bin (Fig.~\ref{fig:robomimic-can}). We adapted this task such that the coke can has two different colors: red or blue. The training data consists of demonstrations from two different demonstrators and the can color is consistent within a single demonstrator. One demonstrator is better than the other one in this task and therefore learning with RGB image can bias the policy to perform poorly on one particular color of can.\ul{The 3 input modalities are proprioception (the position, orientation of the end-effector, and gripper state), RGB image observation from the side and hand cameras, and depth image observation from the side and hand cameras.}. 
MIL learns to mask our the over-specifying RGB modality in this case and achieves a success rate of 56.0\% (see Tab.~\ref{table:succes-rates}).

\medskip

\textbf{Robomimic-Square} is another simulated task from the work of Mandlekar et al.~\cite{mandlekar2021matters}, where a Franka Panda robot is learning to pick up a square nut and put it through the square-shaped pole (Fig.~\ref{fig:robomimic-square}). 
\ul{The 6 input modalities are object state, the position, orientation, angular velocity and linear velocity of the end-effector, and gripper state (open/close)}.
An example run of MIL in this task is shown in Fig.~\ref{fig:robomimic_sample}: as MIL learns to mask out end-effector orientation and velocities, the success rate of the learned policy increases. The average performance of policies learned by MIL and corresponding baselines (across three seeds) are shown in Tab.~\ref{table:succes-rates}. In this task, MIL uses the state-based validation loss. This task is not only long-horizon but also involves precise insertion. Therefore, we further developed a validation loss that leverages a small amount of augmented (failed) trajectory data such that the loss is more selective of high-performing policies instead of policies that get close to the state distribution of validation states but cannot actually finish the task. This version of MIL (MIL-aug) end up finding a policy with the highest the success rate of 71.3\%, achieving 5.6$\times$ performance gain over the vanilla behavioral cloning baseline (BC-NoMask) and is even higher than using the OracleMask. Note that ContinuousMask achieves the lowest performance, which demonstrates that the choice of \textit{binary} mask better addresses the state over-specification problem.

\medskip

\textbf{Bookshelf (real robot)} is a task where a Franka Panda robot needs to learn to reach one of the three sections of the bookshelf, based on the cover of the standing book. As illustrated in Fig.~\ref{fig:realrobot}, \ul{the input modalities are two different RGB viewpoints: one top-down view and one side view}. 
The top-down view can only see the cover of the books laying next to the standing book while the side view can only see the cover of the standing books.
In the training dataset, we introduced a binding between the standing book and the other book in the same section. 
The validation dataset has the same type of binding but between different books. 
In testing scenarios, the standing book and the other book are not paired up. In order to achieve good performance in the test environment, the imitation learning agent needs to mask out the modality with top-down view.
The experiment results in Fig.~\ref{fig:realrobot_results} show the performance of final policies learned by MIL and corresponding baseline methods. MIL recovers the same mask that is picked/designed by demonstrators (the authors) and the learned policy achieves a success rate of 95.2\% (see Tab.~\ref{table:succes-rates}).

%% file: sections/5.conclusion.tex
\section{Conclusion}
\label{sec:conclusion}

\textbf{Summary.} We identify the overfitting issue caused by \textit{state over-specification} in imitation learning from multimodal data and show that we can learn a binary mask through a bi-level optimization algorithm MIL to alleviate the issue. We experiment in both simulated and real-world robotic tasks and demonstrate the effectiveness of our proposed method.

\medskip

\textbf{Limitations.} 
MIL is \textit{computationally expensive} compared to traditional imitation learning methods since it relies on bi-level optimization and the inner loop needs to run imitation learning for every mask update.
In our experiments, we observe that early stopping of the inner loop imitation learning does not hinder the performance of mask learning. Such early stop strategies can be further explored in the future.
In addition, MIL may not return the \textit{smallest} number of modalities, as it may preserve redundant modalities when these modalities do not influence the performance. 
At the same time, the \textit{ordering} of the modalities influences the final mask learned, which means we may not recover the global optimal mask in rare cases in which there exits strong dependency between modalities. Detailed experiments on stability of MIL over modality ordering can be found in the \hyperlink{https://arxiv.org/abs/2209.07682}{Appendix}.


%% file: sections/appendix.tex
\appendix

\subsection{Algorithm Pseudocode}
\begin{algorithm*}[h]
\setstretch{1.15}
\caption{Masked Imitation Learning}\label{alg:mil}
\begin{algorithmic}[1]
\Require training dataset $\mathcal{D}_\text{train}$, validation dataset $\mathcal{D}_\text{val}$, learning rate $\gamma$, convergence error $\epsilon$\;
 \State [Optional] Train forward dynamics model  $\overline{\mathcal{T}}$ with $\mathcal{D}_\text{val}$ and the loss in Eqn. (4)\;
 \State Initialize mask $\Psi \leftarrow \{1\}^M$\;
 \Repeat
    \State $\mathcal{L}_\text{last} = \mathcal{L}_\text{current}$
    \State $\mathcal{L}_\text{current} = inf$
    \For{$i=0,...,(M-1)$}
    \State Set $\Psi[i] = 0$\;
     \State $\Psi^0 \leftarrow \Psi$\;
     \State Set $\Psi[i] = 1$\;
     \State $\Psi^1 \leftarrow \Psi$\;
     \State Initialize policy network $\pi(s;\Psi^0,\theta)$ with the mask using $\Psi^0$\;
    \While{\text{$\mathcal{L}_\text{in}$ not converged}}  
    \State Optimize policy $\pi(s;\Psi^0,\theta)$ with $\mathcal{L}_\text{in}$ on $\mathcal{D}_\text{train}$ using Adam with learning rate $\gamma$\;
    \EndWhile
    \State Compute the validation loss $\mathcal{L}_\text{out}(\Psi^0)$ according to Eqn. (3) or (6) using $\Psi^0$ and $\theta(\Psi^0)$\;
    \State Initialize policy network $\pi(s;\Psi^1,\theta)$ with the mask using $\Psi^1$\;
    \While{\text{$\mathcal{L}_\text{in}$ not converged}}  
    \State Optimize policy $\pi(s;\Psi^0,\theta)$ with $\mathcal{L}_\text{in}$ on $\mathcal{D}_\text{train}$ using Adam with learning rate $\gamma$\;
    \EndWhile
    \State Compute the validation loss $\mathcal{L}_\text{out}(\Psi^1)$ according to Eqn. (3) or (6) using $\Psi^1$ and $\theta(\Psi^1)$\;
   \If{$\mathcal{L}_\text{out}(\Psi^0) < \mathcal{L}_\text{out}(\Psi^1)$}
   \State $\Psi[i] = 0$ \;
   \State $\mathcal{L}_\text{current} = \mathcal{L}_\text{out}(\Psi^0)$\;
   \Else 
   \State $\Psi[i] = 1$\;
   \State $\mathcal{L}_\text{current} = \mathcal{L}_\text{out}(\Psi^1)$\;
   \EndIf
\EndFor      
  \Until{$\lvert \mathcal{L}_\text{current} - \mathcal{L}_\text{last}\rvert \le \epsilon$} 
\State \textbf{return} $\pi_\theta$, $\Psi$
\end{algorithmic}
\end{algorithm*}
We present the pseudocode for the algorithm of Masked Imitation Learning in Algorithm~\ref{alg:mil}, which provides all the details of our implementation of MIL. Line 12-13 and 16-17 are the inner loop imitation learning processes. The outer loop evaluates the validation losses and updates the mask with the one achieving a lower loss. 

\subsection{Generalization of MIL to Stochastic MDP case}
In the stochastic MDP case, the policy we learn is stochastic~\cite{osa2018algorithmic}. Since we use the imitation learning loss for both the inner loss and the outer loss, we just need to replace the losses in Eqn. (2) and (3) with the corresponding imitation learning losses for imitation learning in stochastic MDPs~\cite{fard2011non,ziebart2010modeling,osa2018algorithmic}. 
For the forward model learning in Eqn. (4), we could replace the regression loss with the loss for learning forward model in stochastic MDPs~\cite{antonoglou2021planning}. With the learned dynamics model, for each trajectory $\tau$ in $\mathcal{D}_{\text{val}}$, we generate corresponding trajectories as follows:
\begin{equation}
    \bar{s}_0 = s_0, \quad \bar{s}_{t+1} \sim \overline{\mathcal{T}}(\bar{s}_t, \pi(\bar{s}_t;\Psi,\theta(\Psi))).
\end{equation}
We generate several trajectories $\{\bar{\tau}\}$ for each trajectory in the validation set and use the average distance between the trajectory $\tau$ in the validation set and the trajectories in $\{\bar{\tau}\}$ to account for the stochasticity. 

\subsection{Details of Implementation of Baselines/Ablations}
For the policy network architecture, we use an identity network for the encoder in Reacher setting 1 (targeted distributional shift) and the Robomimic environment since the inputs are all state vectors. For the Reacher setting 2 (with RGB image modality), we use a six-layer convolution network with LeakyRelu as the activation function for the image modality. For the real robot setting, we use resnet18 as the encoder for the image modality. 

For the action predictor, we use a three-layer multilayer perceptron (MLP) with tanh as activation function for Reacher environment and the real robot setting. For the Robomimic environment, we input the masked features into a recurrent neural network (RNN) with LSTM unit. The output of the RNN is fed to a linear layer to output the final action.

For BC-NoMask, we implement the vanilla BC algorithm with all the modalities as input and training the policy network with standard behavior cloning loss.

For MaskDropout, we add a binary mask on the feature of each modality extracted by the corresponding feature encoder to decide whether the modality is used for training the policy. The training process is the same as BC-NoMask except that we randomly change the value of the mask in each update of the network similar to the Dropout technique~\cite{srivastava2014dropout}.

For MaskAverage, we use a policy network with a mask similar to MaskDropout. For the mask, in the Reacher and Robomimic environments, we select five random seeds, which generates five different masks. Then we train a policy for each random mask and use the average testing performance of the policies learned from all the masks. For the real robot setting, since we only have three kinds of masks (use both modalities and use either modality). We average over these three masks.   

For OracleMask, we use a policy network with a fixed mask. Here, the mask is designed with oracle information that is only known to the designer but is not given to training process. In the Reacher environment, we know that the absolute locations and the images are the overfitting modalities for the first and the second settings, so we add a zero mask on these modalities. In the Robomimic environment, we know from the prior work~\cite{mandlekar2021matters} that the end-effector angular and linear velocities are the overfitting modalities, so we add a zero mask on these modalities.

For ContinuousMask, we also use a policy network with a mask but the mask contains continuous value between $[0,1]$. We multiply the mask value with the feature of the corresponding modality. Since the mask is continuous, it is just a new set of learnable variabless that can be updated through standard backpropagation, therefore we directly train the whole network including the encoder, the mask and the action predictor together with the imitation learning loss.

For MIL with online evaluation, we replace the validation loss in our masked imitation learning with an online validation loss. The online validation loss is evaluated by generating rollouts (trajectories of a policy) from interacting with the environment and computing the reward or the success rate of the trajectories. The online validation loss serves as an oracle/upper bound for the validation loss.

\subsection{Details of Experiments in Reacher}
\subsubsection{Setting 1: targeted distributional shift}
The dataset for this setting is gathered using computer-generated optimal demonstrations. For each of the training and validation environments, we trained a Trust Region Policy Optimization (TRPO) agent until reward convergence. Then, using this optimal policy, we rolled out demonstrations in the training and validation environment respectively. Training data consists of targets that are only on the top half of the 2D space, and validation data consists of targets that are only on the bottom half of the 2D space. For this setting, we used 100 training demonstrations and 10 validation demonstrations for all baselines and MIL. Each demonstration trajectory is 50 steps long. The test environment contains targets that are only on the bottom half of the 2D space.

\begin{table}[h]
    \centering
    \addtolength{\tabcolsep}{-2pt}
    \caption{Numerical results of Reacher}
    \label{tab:result_reacher}
    \begin{tabular}{c|cc}
    \toprule
        \multirow{2}{40pt}{Method} & \multicolumn{2}{c}{Reward} \\
        & Setting 1 & Setting 2\\
        \midrule        
        BC-NoMask & -9.156 $\pm$ 0.704 &  -12.167 $\pm$ 0.563\\
        MaskDropout & -9.175 $\pm$ 1.171  & -12.033 $\pm$ 0.149\\
        MaskAverage & -11.28 $\pm$ 1.691 & -10.405 $\pm$ 0.740\\
        ContinuousMask & -8.111 $\pm$ 0.915 & -12.202 $\pm$ 0.631\\
        Redundant-Overfit & -9.587 $\pm$ 0.902 & \textbackslash\\
        Redundant-NonOverfit & -8.963 $\pm$ 2.347 & \textbackslash \\
        RedundntDerivation & -8.807 $\pm$ 0.108 & \textbackslash\\
        MIL & -5.058 $\pm$ 0.299 & -6.063 $\pm$ 0.863 \\
        \midrule
        OracleMask & -5.098 $\pm$ 0.638 & -6.225 $\pm$ 0.691\\
        \bottomrule
    \end{tabular}
\end{table}

We show the numerical results of the Reacher Setting 1 for better comparison between different methods. As shown in Table~\ref{tab:result_reacher}, we observe that MIL outperforms all the baselines statistically significantly, where the largest p-value is $0.013$ with BC-NoMask. Also, the performance of MIL is close to the OracleMask, which indicates that our approach could find a similar mask to the oracle or even detect a better mask.

\subsubsection{Setting 2: standard Reacher plus RGB observation}
The dataset for this setting is gathered using computer-generated optimal demonstrations, through the same process as the targeted distributional shift setting. For this setting, we used the standard Mujoco Reacher environment. We used 20 training demonstrations and 10 validation demonstrations for all baselines and MIL. Each demonstration trajectory is 50 steps long, including one image for each step. 

We show the numerical results of the Reacher Setting 2 for better comparison between different methods. As shown in Table~\ref{tab:result_reacher}, we observe that MIL outperforms all the baselines statistically significantly, where the largest p-value is $0.018$ with MaskAverage.

\begin{figure}[h]
    \centering
        \includegraphics[width=\linewidth]{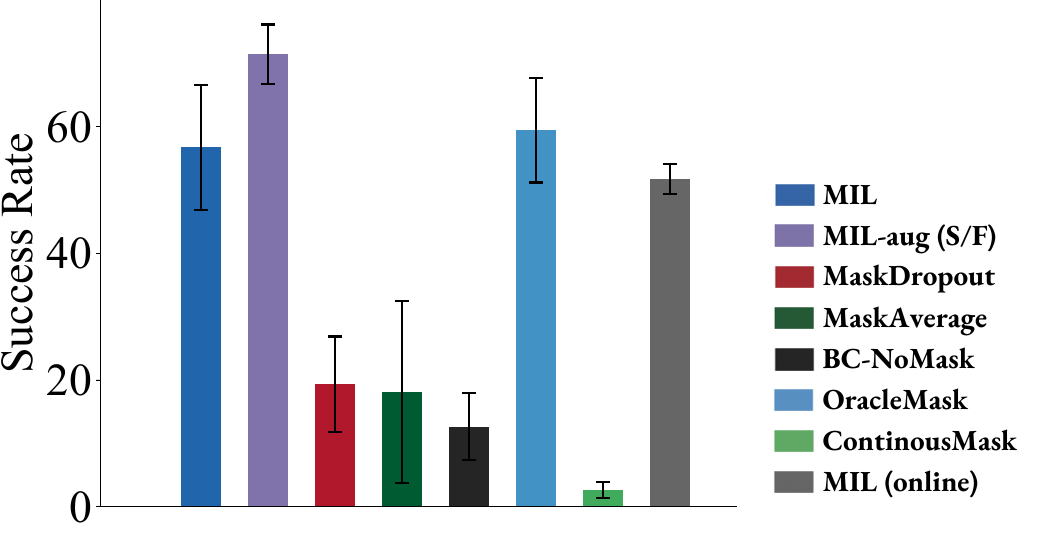}
        \caption{Success rates of policies in \texttt{Robomimic-Square}.}
        \label{fig:robomimic_perf}
\end{figure}

\subsection{Details of Experiments in Robomimic-Square}
\subsubsection{MIL}
We train a forward dynamics model, which consists of a encoder for each modality and also the action, and a next state predictor. No mask is adopted since all the state information is needed to predict the next full state. The encoder transforms the input data of each modality and the action into feature vectors. The state predictor uses the concatenated feature to predict the next state. Here, since both the state and action are vectors, we use multilayer perceptron for encoders and the predictor. We train the forward dynamics model using both training and validation data.

\begin{table}
\centering
    \caption{Numerical results of Robomimic-Square}
    \label{tab:result_robomimic}
    \begin{tabular}{c|c}
    \toprule
        Method & Success Rate(\%) \\
        \midrule        
        BC-NoMask & 12.7$\pm$5.2\\
        MaskDropout & 19.3$\pm$4.5\\
        MaskAverage & 18.1$\pm$14.3\\
        ContinuousMask & 2.7$\pm$1.2\\
        MIL (online) & 51.7$\pm$2.4\\
       MIL-aug (S/F) & 71.3$\pm$4.7\\
        MIL & 56.7$\pm$9.8\\
        \midrule
        OracleMask & 59.3$\pm$8.2\\
        \bottomrule
    \end{tabular}
\end{table}

\subsubsection{MIL-aug (S/F): validation loss using both successful and failed validation trajectories}

To generate augmentation data, we do rollout by setting the start position the same as the initial position of validation data. We add uniform noise $\epsilon \in \mathcal{U}(-\alpha, \alpha)$ to each dimension of the action. Here, $\alpha$ is selected based on the scale of the action. We choose $\alpha = 0.005,0.01,0.05$ to generate $60$ augmented trajectories based on $20$ provided validation data. The rollout length is same as corresponding validation trajectory, and we treat the trajectories that do not succeed within this length as failed trajectories. The $60$ augmented trajectories includes $10$ failed trajectories.

We train the forward dynamics model using training data, validation data, and augmented validation data such that it is accurate near the states covered by these datasets.
During validation, we calculate both the closest success trajectory and the closest failure trajectory using our trained forward model. We represent the loss of closest success trajectory as $\mathcal{L}_{s}$, and represent the loss of closest failure trajectory as $\mathcal{L}_{f}$. To encourage close-to-success trajectories and penalize close-to-failure trajectory, we represent the validation loss as $\mathcal{L}_\text{val}=\exp(\frac{\mathcal{L}_{s}}{t_1}) - \exp(\frac{\mathcal{L}_{f}}{t_2})$. We also want to penalize the trajectories that are far from both success and failure trajectories. We choose $t_1 = 0.2$ and $t_2 = 1$ so that the model whose trajectories are far from both success and failure trajectories is not getting smallest loss. We show the numerical results of the success rate in Table~\ref{tab:result_robomimic}.

\subsection{Details of Experiments in Robomimic-Can}
We train a forward dynamics model, which consists of a encoder for end effecftor proprioception modality and also the action, and a next state predictor. No mask is adopted since all the state information is needed to predict the next full state. The encoder transforms the input data of each modality and the action into feature vectors. The state predictor uses the concatenated feature to predict the next state. Here, since both the state and action are vectors, we use multilayer perceptron for encoders and the predictor. We train the forward dynamics model using both training and validation data.

\begin{figure}[h]
  \centering
    \includegraphics[width=0.4\textwidth]{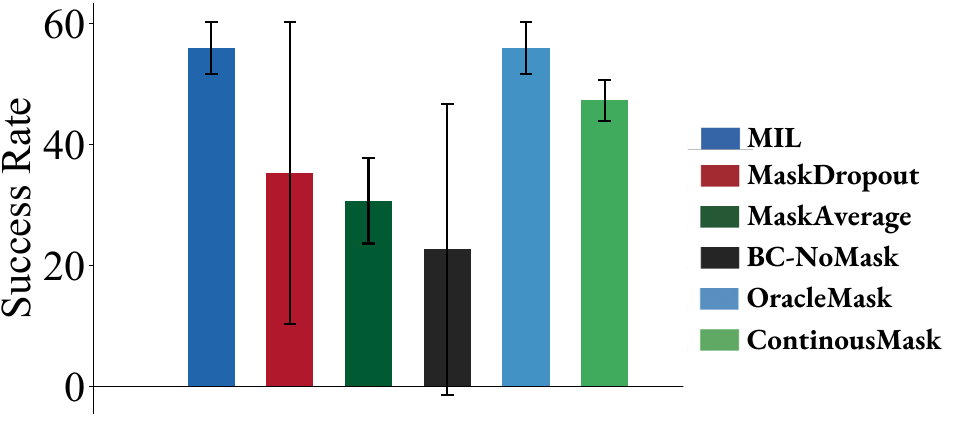}
  \caption{\small{Success rates of policies in \texttt{Robomimic-Can}.}}
  \label{fig:realrobot_results}
  \vspace{-0.4cm}
\end{figure}

\begin{table}
    \centering
    \caption{Numerical results of Real Robot}
    \label{tab:result_real_robot}
    \begin{tabular}{c|c}
    \toprule
        Method & Success Rate(\%) \\
        \midrule        
        BC-NoMask & 54.17\\
        MaskDropout & 47.9\\
        MaskAverage & 60.9\\
        ContinuousMask & 19.8\\
        MIL & 95.24\\
        \midrule
        OracleMask & 95.24\\
        \bottomrule
    \end{tabular}
\end{table}

\begin{figure}[ht]
    \centering
    \begin{subfigure}[b]{0.5\textwidth}
         \centering
         \includegraphics[width=\textwidth]{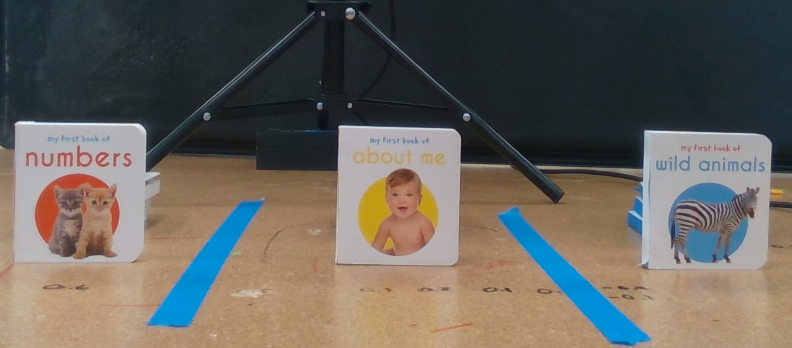}
         \caption{\texttt{Front view}}
         \label{fig:front_view}
     \end{subfigure}
    \begin{subfigure}[b]{0.5\textwidth}
         \centering
         \includegraphics[width=\textwidth]{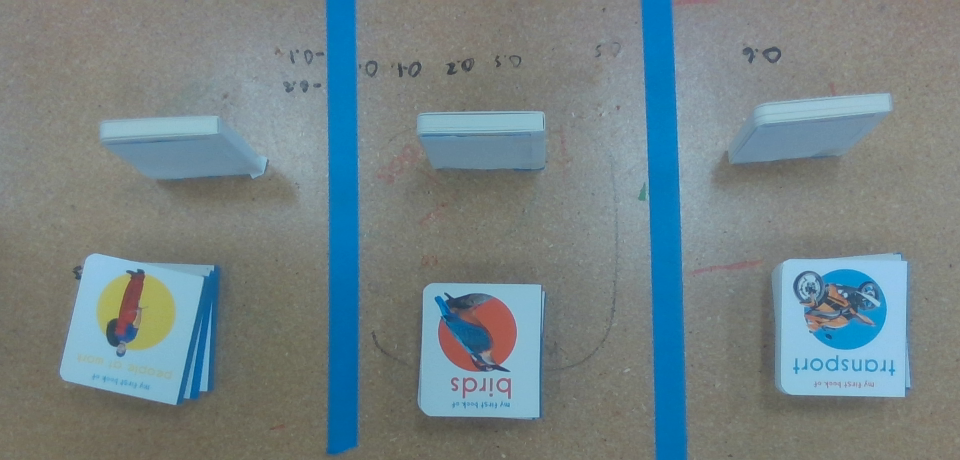}
         \caption{\texttt{Top view}}
         \label{fig:top_view}
     \end{subfigure}
    \caption{Visualization of two views of experiments in Bookshelf (real robot).}
    \label{fig:views}
\end{figure}

\begin{figure}[h]
  \centering
    \includegraphics[width=0.4\textwidth]{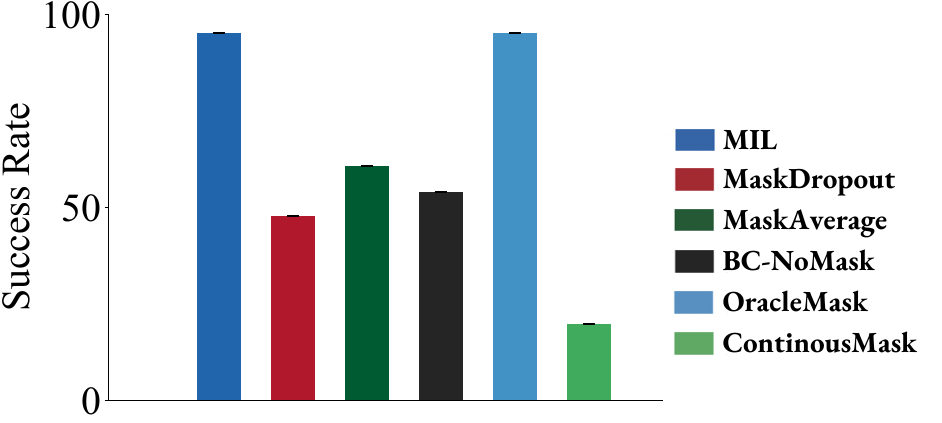}
  \caption{\small{Success rates of policies in real robot task \texttt{Bookshelf}.}}
  \label{fig:realrobot_results}
  \vspace{-0.4cm}
\end{figure}

\begin{figure}[h]
    \centering
    \includegraphics[width=\linewidth]{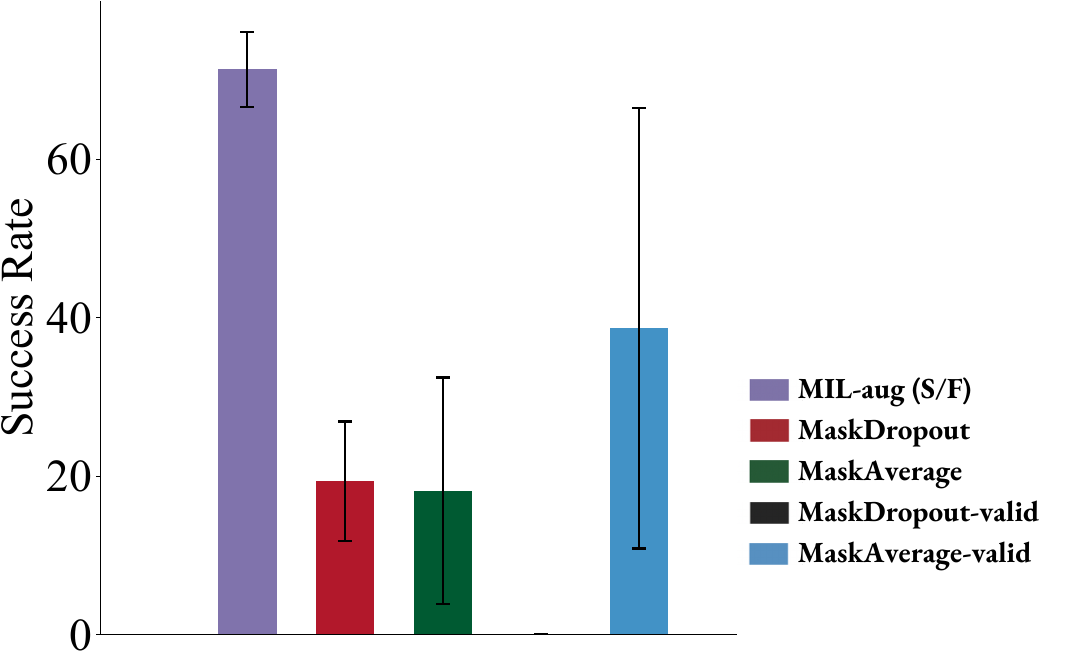}
    \caption{Experimental results comparison of using validation data to select policy in \texttt{Robomimic-Square} }
    \label{fig:robomimic_select_using_valid_data}
    \vspace{-0.3cm}
\end{figure}

\begin{figure}[h]
    \centering
    \includegraphics[width=\linewidth]{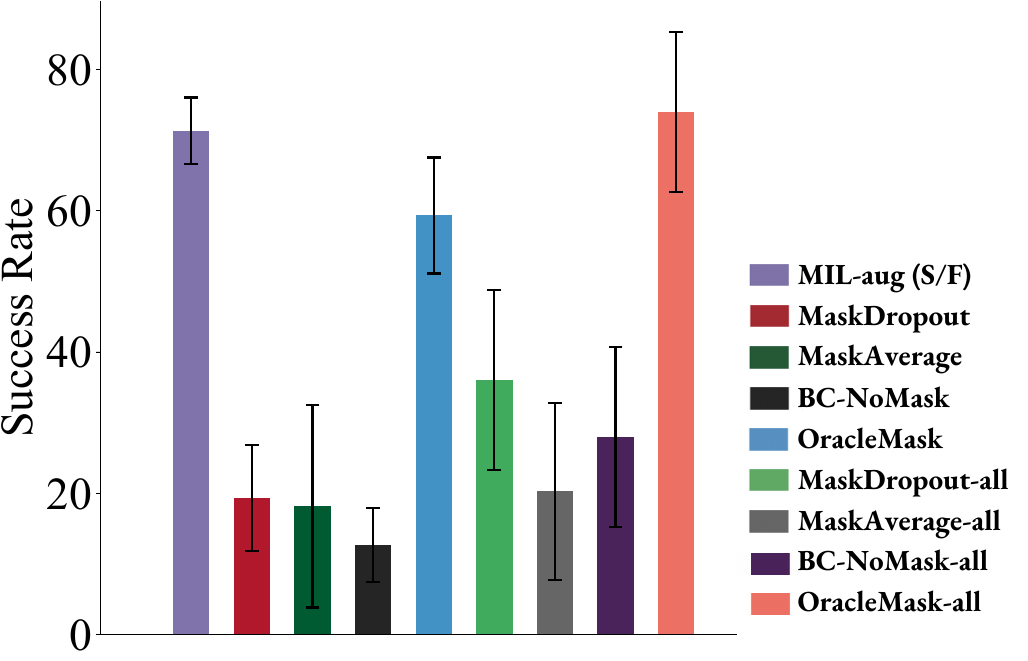}
    \caption{Experimental results comparison of including validation data in training in \texttt{Robomimic-Square} }
    \label{fig:robomimic_train_alldata}
    \vspace{-0.3cm}
\end{figure}
\subsection{Details of Experiments in Bookshelf (real robot)}

Figure~\ref{fig:views} shows the visualization of two views of the real robot bookshelf experiments. The task for the robot arm is to go to the subsection after the standing orange book. In the front view, there are 3 standing books with different colors, we cannot see the books behind them. In the top view, there are 3 stack of books with different front cover color, since the back cover and pages are white, we cannot distinguish the three standing books.

We collected both training and validation data by human controlling the joystick. During training, we train with collected data with a fix “standing book-book stack” combination: yellow-orange, orange-blue, blue-yellow. The blue book stack is always behind the standing orange book, the robot always goes to find the blue book stack. During validation, we use collected data with different “standing book-book stack” combination, so the standing orange book no longer corresponds to the blue book stack. During testing, we test with images from real-time setting, and we tested with all different front cover combinations. We show the numerical results of the success rate in Table~\ref{tab:result_real_robot}.

\begin{table}[h]
    \centering
    \addtolength{\tabcolsep}{-3pt}
    \caption{Varying Modality Order for \texttt{Robomimic-Square}}
    \label{tab:change_order}
    \begin{tabular}{c|c|c|c|c}
    \toprule
        Seed & Order & Learned Mask & Success Rate & Average \\
        \midrule        
        \multirow{4}{*}{\#1} & 1 &  110101 & 54\% & \multirow{4}{*}{50.0 $\pm$ 7.8}\\
         & 2 &  100011 & 48\% & \\
         & 3 &  101011 & 58\% & \\
         & 4 &  110110 & 40\% & \\
        \midrule
        \multirow{4}{*}{\#2} & 1 &  110000 & 74\% & \multirow{4}{*}{71.0 $\pm$ 6.0}\\
         & 2 &  110011 & 74\% & \\
         & 3 &  111000 & 62\% & \\
         & 4 &  110011 & 74\% & \\
        \bottomrule
    \end{tabular}
\end{table}

\subsection{Additional Experiments in Robomimic-Square}

\subsubsection{Changing modality order}
Since MIL learns the mask bit by bit, to ensure the stability of our algorithm, we also tested our algorithms with different modality ordering. Table~\ref{tab:change_order} shows the result of changing modality order experiments on MIL-aug. We randomly selected four different modality orders for each seed and tested for two seeds. We then transferred the ordering back and show the learned masks in the table. From the table we can see, all learned masks have successfully learned the first bit as 1, which represents object information and is the most crucial bit in this task. In addition, for all four different modality orders, MIL-aug can learn masks with similar success rates. Thus, the stability of the discovered policy performance maintains when modality ordering changes. However, we also note that there isn't dependency between modalities in this experiment and when there exist heavy dependency between modalities, MIL may not recover the optimal mask.

\begin{table}[h]
\centering
\addtolength{\tabcolsep}{-2pt}
\caption{Continuous Mask for \texttt{Robomimic-Square}}
\label{tab:continuous}
\begin{tabular}{c|c|c}
\hline
\multirow{2}{*}{\begin{tabular}[c]{@{}c@{}}Extra Modality \\ Weight\end{tabular}} & \multirow{2}{*}{\begin{tabular}[c]{@{}c@{}}Success Rate(\%) \\ of Epoch 500\end{tabular}} & \multirow{2}{*}{\begin{tabular}[c]{@{}c@{}}Best Success Rate(\%) \\ over 2000 Epochs\end{tabular}} \\
&   &  \\ \hline
0 & 18 & 70 \\ \hline
0.03 & 6 & 52  \\ \hline
0.06 & 8 & 40 \\ \hline
0.09 & 4  & 40     \\ \hline
0.12 & 0  & 24    \\ \hline
0.15 & 0 & 24 \\ \hline
\end{tabular}
\end{table}

\subsubsection{Selecting Masks based on validation information}
Since in MIL, we use validation data $\mathcal{D}_\text{val}$ to select between masks, we also included results of using $\mathcal{D}_\text{val}$ in a similar way for baselines MaskDropout and MaskAverage. For MaskDropout, we uses the trained policy, then evaluate it with validation data with differrent masks(we pass all possible $2^6$ masks), and select the mask that has the lowest validation loss. Fig. \ref{fig:robomimic_select_using_valid_data} shows the results. This method does not always choose the best mask. In all three seeds of MaskDropout policy, the selected mask has 0\% success rate. For MaskAverage, we take the group of trained policies with fixed random masks, and select the best policy with lowest validation loss from the group. This method does not always select the best mask from the group, and the performance is capped by the randomly selected masks. From Fig. \ref{fig:robomimic_select_using_valid_data} we can observe, MaskAverage-valid has 38.67\% success rate, which is still much lower than MIL-aug(71.3\%).

\subsubsection{Baselines training with all data}
We included the experiment of training baselines with both $\mathcal{D}_\text{train}$ and $\mathcal{D}_\text{val}$ and included the results in Fig. \ref{fig:robomimic_train_alldata}. The figure shows that although with more training data the success rates increase for all baselines, MIL-aug still outperform all baselines to a great extent(except oracle, we include oracle result to show the natural performance increase by including more training data)

\begin{figure}[h]
    \centering
    \includegraphics[width=\linewidth]{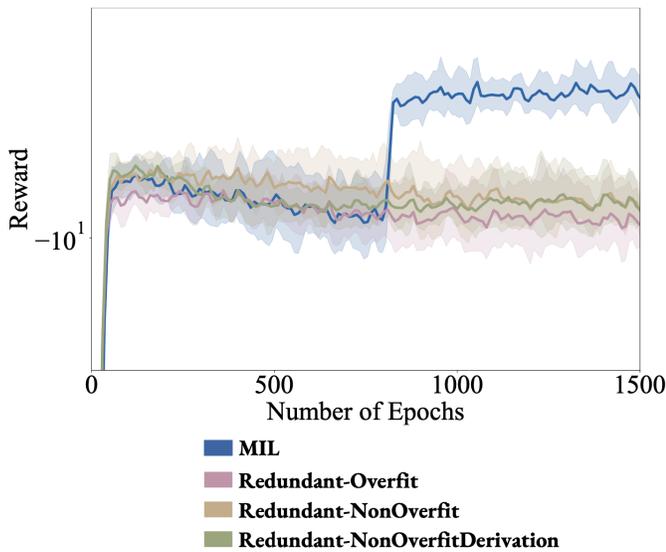}
    \label{fig:mod8_redundant}
    \caption{Experimental results of distributionally Shifted Goals with redundant modality information}
    \vspace{-0.3cm}
\end{figure}

\subsubsection{Continuous Mask}
We also experimented with continuous mask with fixed mask weights. In \texttt{Robomimic-Square}, since there are in total six modalities, and two of them are extra modalities since they cause performance drop if added. In this experiment, we fix the mask weight of two extra modalities and gradually increase this mask weight. When extra modality weight is 0, each of the other four modalities has 0.25 weight. When extra modality weight is 0.03, this means both two extra modalities has 0.03 weight, each of the other four modalities has 0.235 weight. Table~\ref{tab:continuous} shows the result. We tested extra modality weight from 0 to 0.15, since 0.166 would be uniformly weighting the six modalities. We recorded both epoch 500 success rate, and the best success rate over all 2000 epochs. From the result, we can observe that the performance drops drastically even if we only added 0.03 each time to the mask weight. This experiment shows that even very small amount of extra modalities could affect the performance seriously. Thus, our method focuses on learning binary masks instead of continuous masks.

%% file: root.bbl
\begin{thebibliography}{54}
\providecommand{\natexlab}[1]{#1}
\providecommand{\url}[1]{#1}
\csname url@samestyle\endcsname
\providecommand{\newblock}{\relax}
\providecommand{\bibinfo}[2]{#2}
\providecommand{\BIBentrySTDinterwordspacing}{\spaceskip=0pt\relax}
\providecommand{\BIBentryALTinterwordstretchfactor}{4}
\providecommand{\BIBentryALTinterwordspacing}{\spaceskip=\fontdimen2\font plus
\BIBentryALTinterwordstretchfactor\fontdimen3\font minus
  \fontdimen4\font\relax}
\providecommand{\BIBforeignlanguage}[2]{{%
\expandafter\ifx\csname l@#1\endcsname\relax
\typeout{** WARNING: IEEEtranN.bst: No hyphenation pattern has been}%
\typeout{** loaded for the language `#1'. Using the pattern for}%
\typeout{** the default language instead.}%
\else
\language=\csname l@#1\endcsname
\fi
#2}}
\providecommand{\BIBdecl}{\relax}
\BIBdecl

\bibitem[Ernst and B{\"u}lthoff(2004)]{ernst2004merging}
M.~O. Ernst and H.~H. B{\"u}lthoff, ``Merging the senses into a robust
  percept,'' \emph{Trends in cognitive sciences}, vol.~8, no.~4, pp. 162--169,
  2004.

\bibitem[Stein and Meredith(1993)]{stein1993merging}
B.~E. Stein and M.~A. Meredith, \emph{The merging of the senses.}\hskip 1em
  plus 0.5em minus 0.4em\relax The MIT press, 1993.

\bibitem[Calandra et~al.(2018)Calandra, Owens, Jayaraman, Lin, Yuan, Malik,
  Adelson, and Levine]{calandra2018more}
R.~Calandra, A.~Owens, D.~Jayaraman, J.~Lin, W.~Yuan, J.~Malik, E.~H. Adelson,
  and S.~Levine, ``More than a feeling: Learning to grasp and regrasp using
  vision and touch,'' \emph{IEEE Robotics and Automation Letters}, vol.~3,
  no.~4, pp. 3300--3307, 2018.

\bibitem[Zhang et~al.(2019)Zhang, Sharma, Veloso, and
  Kroemer]{zhang2019leveraging}
K.~Zhang, M.~Sharma, M.~Veloso, and O.~Kroemer, ``Leveraging multimodal haptic
  sensory data for robust cutting,'' in \emph{2019 IEEE-RAS 19th International
  Conference on Humanoid Robots (Humanoids)}.\hskip 1em plus 0.5em minus
  0.4em\relax IEEE, 2019, pp. 409--416.

\bibitem[Mu et~al.(2021)Mu, Ling, Xiang, Yang, Li, Tao, Huang, Jia, and
  Su]{mu2021maniskill}
T.~Mu, Z.~Ling, F.~Xiang, D.~Yang, X.~Li, S.~Tao, Z.~Huang, Z.~Jia, and H.~Su,
  ``Maniskill: Generalizable manipulation skill benchmark with large-scale
  demonstrations,'' \emph{arXiv preprint arXiv:2107.14483}, 2021.

\bibitem[Liu et~al.(2019{\natexlab{a}})Liu, Liu, Li, Chen, Zhang, and
  Liu]{liu2019flexible}
T.~Liu, H.~Liu, Y.-F. Li, Z.~Chen, Z.~Zhang, and S.~Liu, ``Flexible ftir
  spectral imaging enhancement for industrial robot infrared vision sensing,''
  \emph{IEEE Transactions on Industrial Informatics}, vol.~16, no.~1, pp.
  544--554, 2019.

\bibitem[Abad and Ranasinghe(2020)]{abad2020visuotactile}
A.~C. Abad and A.~Ranasinghe, ``Visuotactile sensors with emphasis on gelsight
  sensor: A review,'' \emph{IEEE Sensors Journal}, vol.~20, no.~14, pp.
  7628--7638, 2020.

\bibitem[Dai et~al.(2022)Dai, Wang, Rojas, Harber, Tian, Paiva, Gnehm,
  Schindewolf, Choset, Webster-Wood, et~al.]{dai2022design}
K.~Dai, X.~Wang, A.~M. Rojas, E.~Harber, Y.~Tian, N.~Paiva, J.~Gnehm,
  E.~Schindewolf, H.~Choset, V.~A. Webster-Wood \emph{et~al.}, ``Design of a
  biomimetic tactile sensor for material classification,'' \emph{2022 IEEE
  International Conference on Robotics and Automation (ICRA)}, 2022.

\bibitem[Mandlekar et~al.(2021)Mandlekar, Xu, Wong, Nasiriany, Wang, Kulkarni,
  Fei-Fei, Savarese, Zhu, and Mart{\'\i}n-Mart{\'\i}n]{mandlekar2021matters}
A.~Mandlekar, D.~Xu, J.~Wong, S.~Nasiriany, C.~Wang, R.~Kulkarni, L.~Fei-Fei,
  S.~Savarese, Y.~Zhu, and R.~Mart{\'\i}n-Mart{\'\i}n, ``What matters in
  learning from offline human demonstrations for robot manipulation,''
  \emph{arXiv preprint arXiv:2108.03298}, 2021.

\bibitem[Xiao et~al.(2020)Xiao, Codevilla, Gurram, Urfalioglu, and
  L{\'o}pez]{xiao2020multimodal}
Y.~Xiao, F.~Codevilla, A.~Gurram, O.~Urfalioglu, and A.~M. L{\'o}pez,
  ``Multimodal end-to-end autonomous driving,'' \emph{IEEE Transactions on
  Intelligent Transportation Systems}, 2020.

\bibitem[Tomar et~al.(2021)Tomar, Zhang, Calandra, Taylor, and
  Pineau]{tomar2021model}
M.~Tomar, A.~Zhang, R.~Calandra, M.~E. Taylor, and J.~Pineau, ``Model-invariant
  state abstractions for model-based reinforcement learning,'' in
  \emph{Self-Supervision for Reinforcement Learning Workshop-ICLR 2021}, 2021.

\bibitem[Akinola et~al.(2020)Akinola, Varley, and
  Kalashnikov]{akinola2020learning}
I.~Akinola, J.~Varley, and D.~Kalashnikov, ``Learning precise 3d manipulation
  from multiple uncalibrated cameras,'' in \emph{2020 IEEE International
  Conference on Robotics and Automation (ICRA)}.\hskip 1em plus 0.5em minus
  0.4em\relax IEEE, 2020, pp. 4616--4622.

\bibitem[Liu et~al.(2017)Liu, Siravuru, Prabhakar, Veloso, and
  Kantor]{liu2017learning}
G.-H. Liu, A.~Siravuru, S.~Prabhakar, M.~Veloso, and G.~Kantor, ``Learning
  end-to-end multimodal sensor policies for autonomous navigation,'' in
  \emph{Conference on Robot Learning}.\hskip 1em plus 0.5em minus 0.4em\relax
  PMLR, 2017, pp. 249--261.

\bibitem[Lee et~al.(2020)Lee, Zhu, Zachares, Tan, Srinivasan, Savarese,
  Fei-Fei, Garg, and Bohg]{lee2020making}
M.~A. Lee, Y.~Zhu, P.~Zachares, M.~Tan, K.~Srinivasan, S.~Savarese, L.~Fei-Fei,
  A.~Garg, and J.~Bohg, ``Making sense of vision and touch: Learning multimodal
  representations for contact-rich tasks,'' \emph{IEEE Transactions on
  Robotics}, vol.~36, no.~3, pp. 582--596, 2020.

\bibitem[Lee et~al.(2021)Lee, Tan, Zhu, and Bohg]{lee2021detect}
M.~A. Lee, M.~Tan, Y.~Zhu, and J.~Bohg, ``Detect, reject, correct: Crossmodal
  compensation of corrupted sensors,'' in \emph{2021 IEEE International
  Conference on Robotics and Automation (ICRA)}.\hskip 1em plus 0.5em minus
  0.4em\relax IEEE, 2021, pp. 909--916.

\bibitem[Gao et~al.(2016)Gao, Hendricks, Kuchenbecker, and
  Darrell]{gao2016deep}
Y.~Gao, L.~A. Hendricks, K.~J. Kuchenbecker, and T.~Darrell, ``Deep learning
  for tactile understanding from visual and haptic data,'' in \emph{2016 IEEE
  International Conference on Robotics and Automation (ICRA)}.\hskip 1em plus
  0.5em minus 0.4em\relax IEEE, 2016, pp. 536--543.

\bibitem[Bekiroglu et~al.(2011)Bekiroglu, Detry, and
  Kragic]{bekiroglu2011learning}
Y.~Bekiroglu, R.~Detry, and D.~Kragic, ``Learning tactile characterizations of
  object-and pose-specific grasps,'' in \emph{2011 IEEE/RSJ international
  conference on Intelligent Robots and Systems}.\hskip 1em plus 0.5em minus
  0.4em\relax IEEE, 2011, pp. 1554--1560.

\bibitem[Sinapov et~al.(2014)Sinapov, Schenck, and
  Stoytchev]{sinapov2014learning}
J.~Sinapov, C.~Schenck, and A.~Stoytchev, ``Learning relational object
  categories using behavioral exploration and multimodal perception,'' in
  \emph{2014 IEEE international conference on robotics and automation
  (ICRA)}.\hskip 1em plus 0.5em minus 0.4em\relax IEEE, 2014, pp. 5691--5698.

\bibitem[Sung et~al.(2017)Sung, Salisbury, and Saxena]{sung2017learning}
J.~Sung, J.~K. Salisbury, and A.~Saxena, ``Learning to represent haptic
  feedback for partially-observable tasks,'' in \emph{2017 IEEE International
  Conference on Robotics and Automation (ICRA)}.\hskip 1em plus 0.5em minus
  0.4em\relax IEEE, 2017, pp. 2802--2809.

\bibitem[Du et~al.(2022)Du, Lee, Nair, and Finn]{du2022play}
M.~Du, O.~Y. Lee, S.~Nair, and C.~Finn, ``Play it by ear: Learning skills
  amidst occlusion through audio-visual imitation learning,'' \emph{Robotics:
  Science and Systems}, 2022.

\bibitem[Dean et~al.(2020)Dean, Tulsiani, and Gupta]{dean2020see}
V.~Dean, S.~Tulsiani, and A.~Gupta, ``See, hear, explore: Curiosity via
  audio-visual association,'' \emph{Advances in Neural Information Processing
  Systems}, vol.~33, pp. 14\,961--14\,972, 2020.

\bibitem[Yang et~al.(2017)Yang, Ramesh, Chitta, Madhvanath, Bernal, and
  Luo]{yang2017deep}
X.~Yang, P.~Ramesh, R.~Chitta, S.~Madhvanath, E.~A. Bernal, and J.~Luo, ``Deep
  multimodal representation learning from temporal data,'' in \emph{Proceedings
  of the IEEE conference on computer vision and pattern recognition}, 2017, pp.
  5447--5455.

\bibitem[Chen et~al.(2021)Chen, Lee, and Soh]{chen2021multi}
K.~Chen, Y.~Lee, and H.~Soh, ``Multi-modal mutual information (mummi) training
  for robust self-supervised deep reinforcement learning,'' in \emph{2021 IEEE
  International Conference on Robotics and Automation (ICRA)}.\hskip 1em plus
  0.5em minus 0.4em\relax IEEE, 2021, pp. 4274--4280.

\bibitem[Argall et~al.(2009)Argall, Chernova, Veloso, and
  Browning]{argall2009survey}
B.~D. Argall, S.~Chernova, M.~Veloso, and B.~Browning, ``A survey of robot
  learning from demonstration,'' \emph{Robotics and autonomous systems},
  vol.~57, no.~5, pp. 469--483, 2009.

\bibitem[Osa et~al.(2018)Osa, Pajarinen, Neumann, Bagnell, Abbeel, Peters,
  et~al.]{osa2018algorithmic}
T.~Osa, J.~Pajarinen, G.~Neumann, J.~A. Bagnell, P.~Abbeel, J.~Peters
  \emph{et~al.}, ``An algorithmic perspective on imitation learning,''
  \emph{Foundations and Trends{\textregistered} in Robotics}, vol.~7, no. 1-2,
  pp. 1--179, 2018.

\bibitem[Bain and Sammut(1995)]{bain1995framework}
M.~Bain and C.~Sammut, ``A framework for behavioural cloning.'' in
  \emph{Machine Intelligence 15}, 1995.

\bibitem[Schroecker and Isbell(2017)]{schroecker2017state}
Y.~Schroecker and C.~L. Isbell, ``State aware imitation learning,'' in
  \emph{NeurIPS}, 2017.

\bibitem[Torabi et~al.(2019)Torabi, Warnell, and Stone]{torabi2018generative}
F.~Torabi, G.~Warnell, and P.~Stone, ``Generative adversarial imitation from
  observation,'' \emph{Imitation, Intent, and Interaction (I3) Workshop at
  ICML}, 2019.

\bibitem[Sun et~al.(2019)Sun, Vemula, Boots, and Bagnell]{sun2019provably}
W.~Sun, A.~Vemula, B.~Boots, and D.~Bagnell, ``Provably efficient imitation
  learning from observation alone,'' in \emph{ICML}, 2019.

\bibitem[Fu et~al.(2018)Fu, Luo, and Levine]{fu2017learning}
J.~Fu, K.~Luo, and S.~Levine, ``Learning robust rewards with adverserial
  inverse reinforcement learning,'' in \emph{ICLR}, 2018.

\bibitem[Liu et~al.(2019{\natexlab{b}})Liu, Ling, Mu, and Su]{liu2019state}
F.~Liu, Z.~Ling, T.~Mu, and H.~Su, ``State alignment-based imitation
  learning,'' in \emph{ICLR}, 2019.

\bibitem[Cao et~al.(2021)Cao, Hao, Li, and Sadigh]{cao2021corl}
Z.~Cao, Y.~Hao, M.~Li, and D.~Sadigh, ``Learning feasibility to imitate
  demonstrators with different dynamics,'' in \emph{Conference on Robot
  Learning}, 2021.

\bibitem[Zhang et~al.(2021)Zhang, Bengio, Hardt, Recht, and
  Vinyals]{zhang2021understanding}
C.~Zhang, S.~Bengio, M.~Hardt, B.~Recht, and O.~Vinyals, ``Understanding deep
  learning (still) requires rethinking generalization,'' \emph{Communications
  of the ACM}, vol.~64, no.~3, pp. 107--115, 2021.

\bibitem[Arpit et~al.(2017)Arpit, Jastrz{\k{e}}bski, Ballas, Krueger, Bengio,
  Kanwal, Maharaj, Fischer, Courville, Bengio, et~al.]{arpit2017closer}
D.~Arpit, S.~Jastrz{\k{e}}bski, N.~Ballas, D.~Krueger, E.~Bengio, M.~S. Kanwal,
  T.~Maharaj, A.~Fischer, A.~Courville, Y.~Bengio \emph{et~al.}, ``A closer
  look at memorization in deep networks,'' in \emph{International conference on
  machine learning}.\hskip 1em plus 0.5em minus 0.4em\relax PMLR, 2017, pp.
  233--242.

\bibitem[Arjovsky et~al.(2019)Arjovsky, Bottou, Gulrajani, and
  Lopez-Paz]{arjovsky2019invariant}
M.~Arjovsky, L.~Bottou, I.~Gulrajani, and D.~Lopez-Paz, ``Invariant risk
  minimization,'' \emph{arXiv preprint arXiv:1907.02893}, 2019.

\bibitem[Mahajan et~al.(2021)Mahajan, Tople, and Sharma]{mahajan2021domain}
D.~Mahajan, S.~Tople, and A.~Sharma, ``Domain generalization using causal
  matching,'' in \emph{International Conference on Machine Learning}.\hskip 1em
  plus 0.5em minus 0.4em\relax PMLR, 2021, pp. 7313--7324.

\bibitem[Chen et~al.(2020)Chen, Wei, Kumar, and Ma]{chen2020self}
Y.~Chen, C.~Wei, A.~Kumar, and T.~Ma, ``Self-training avoids using spurious
  features under domain shift,'' \emph{Advances in Neural Information
  Processing Systems}, vol.~33, pp. 21\,061--21\,071, 2020.

\bibitem[Xie et~al.(2020)Xie, Luong, Hovy, and Le]{xie2020self}
Q.~Xie, M.-T. Luong, E.~Hovy, and Q.~V. Le, ``Self-training with noisy student
  improves imagenet classification,'' in \emph{Proceedings of the IEEE/CVF
  conference on computer vision and pattern recognition}, 2020, pp.
  10\,687--10\,698.

\bibitem[Srivastava et~al.(2014)Srivastava, Hinton, Krizhevsky, Sutskever, and
  Salakhutdinov]{srivastava2014dropout}
N.~Srivastava, G.~Hinton, A.~Krizhevsky, I.~Sutskever, and R.~Salakhutdinov,
  ``Dropout: a simple way to prevent neural networks from overfitting,''
  \emph{The journal of machine learning research}, vol.~15, no.~1, pp.
  1929--1958, 2014.

\bibitem[Yang and Pedersen(1997)]{yang1997comparative}
Y.~Yang and J.~O. Pedersen, ``A comparative study on feature selection in text
  categorization,'' in \emph{Icml}, vol.~97, no. 412-420.\hskip 1em plus 0.5em
  minus 0.4em\relax Nashville, TN, USA, 1997, p.~35.

\bibitem[Bach(2008)]{bach2008bolasso}
F.~R. Bach, ``Bolasso: model consistent lasso estimation through the
  bootstrap,'' in \emph{Proceedings of the 25th international conference on
  Machine learning}, 2008, pp. 33--40.

\bibitem[Urbanowicz et~al.(2018)Urbanowicz, Meeker, La~Cava, Olson, and
  Moore]{urbanowicz2018relief}
R.~J. Urbanowicz, M.~Meeker, W.~La~Cava, R.~S. Olson, and J.~H. Moore,
  ``Relief-based feature selection: Introduction and review,'' \emph{Journal of
  biomedical informatics}, vol.~85, pp. 189--203, 2018.

\bibitem[Pacelli and Majumdar(2020)]{pacelli2020learning}
V.~Pacelli and A.~Majumdar, ``Learning task-driven control policies via
  information bottlenecks,'' \emph{arXiv preprint arXiv:2002.01428}, 2020.

\bibitem[Lu et~al.(2020)Lu, Lee, Abbeel, and Tiomkin]{lu2020dynamics}
X.~Lu, K.~Lee, P.~Abbeel, and S.~Tiomkin, ``Dynamics generalization via
  information bottleneck in deep reinforcement learning,'' \emph{arXiv preprint
  arXiv:2008.00614}, 2020.

\bibitem[Ahuja et~al.(2020)Ahuja, Shanmugam, Varshney, and
  Dhurandhar]{ahuja2020invariant}
K.~Ahuja, K.~Shanmugam, K.~Varshney, and A.~Dhurandhar, ``Invariant risk
  minimization games,'' in \emph{International Conference on Machine
  Learning}.\hskip 1em plus 0.5em minus 0.4em\relax PMLR, 2020, pp. 145--155.

\bibitem[Tobin et~al.(2017)Tobin, Fong, Ray, Schneider, Zaremba, and
  Abbeel]{tobin2017domain}
J.~Tobin, R.~Fong, A.~Ray, J.~Schneider, W.~Zaremba, and P.~Abbeel, ``Domain
  randomization for transferring deep neural networks from simulation to the
  real world,'' in \emph{2017 IEEE/RSJ international conference on intelligent
  robots and systems (IROS)}.\hskip 1em plus 0.5em minus 0.4em\relax IEEE,
  2017, pp. 23--30.

\bibitem[Peng et~al.(2018)Peng, Andrychowicz, Zaremba, and Abbeel]{peng2018sim}
X.~B. Peng, M.~Andrychowicz, W.~Zaremba, and P.~Abbeel, ``Sim-to-real transfer
  of robotic control with dynamics randomization,'' in \emph{2018 IEEE
  international conference on robotics and automation (ICRA)}.\hskip 1em plus
  0.5em minus 0.4em\relax IEEE, 2018, pp. 3803--3810.

\bibitem[Bousmalis et~al.(2018)Bousmalis, Irpan, Wohlhart, Bai, Kelcey,
  Kalakrishnan, Downs, Ibarz, Pastor, Konolige, et~al.]{bousmalis2018using}
K.~Bousmalis, A.~Irpan, P.~Wohlhart, Y.~Bai, M.~Kelcey, M.~Kalakrishnan,
  L.~Downs, J.~Ibarz, P.~Pastor, K.~Konolige \emph{et~al.}, ``Using simulation
  and domain adaptation to improve efficiency of deep robotic grasping,'' in
  \emph{2018 IEEE international conference on robotics and automation
  (ICRA)}.\hskip 1em plus 0.5em minus 0.4em\relax IEEE, 2018, pp. 4243--4250.

\bibitem[Wright(2015)]{wright2015coordinate}
S.~J. Wright, ``Coordinate descent algorithms,'' \emph{Mathematical
  Programming}, vol. 151, no.~1, pp. 3--34, 2015.

\bibitem[Ross et~al.(2011)Ross, Gordon, and Bagnell]{ross2011reduction}
S.~Ross, G.~Gordon, and D.~Bagnell, ``A reduction of imitation learning and
  structured prediction to no-regret online learning,'' in \emph{AISTATS},
  2011.

\bibitem[Ross and Bagnell(2010)]{ross2010efficient}
S.~Ross and D.~Bagnell, ``Efficient reductions for imitation learning,'' in
  \emph{AISTATS}, 2010.

\bibitem[Fard and Pineau(2011)]{fard2011non}
M.~M. Fard and J.~Pineau, ``Non-deterministic policies in markovian decision
  processes,'' \emph{Journal of Artificial Intelligence Research}, vol.~40, pp.
  1--24, 2011.

\bibitem[Ziebart(2010)]{ziebart2010modeling}
B.~D. Ziebart, \emph{Modeling purposeful adaptive behavior with the principle
  of maximum causal entropy}.\hskip 1em plus 0.5em minus 0.4em\relax Carnegie
  Mellon University, 2010.

\bibitem[Antonoglou et~al.(2021)Antonoglou, Schrittwieser, Ozair, Hubert, and
  Silver]{antonoglou2021planning}
I.~Antonoglou, J.~Schrittwieser, S.~Ozair, T.~K. Hubert, and D.~Silver,
  ``Planning in stochastic environments with a learned model,'' in
  \emph{International Conference on Learning Representations}, 2021.

\end{thebibliography}
